\documentclass[journal]{IEEEtran} 
\IEEEoverridecommandlockouts
\usepackage{booktabs} 
\usepackage{tabularx} 
\usepackage{xcolor} 
\usepackage{graphicx} 
\usepackage{cite}
\usepackage{amsmath,amssymb,amsfonts,amsthm}
\usepackage{threeparttable}
\usepackage{graphicx}
\usepackage{textcomp}{\tiny }
\usepackage{multirow}
\usepackage{xcolor}
\usepackage{bm,color}
\usepackage{lmodern} 
\usepackage{makecell}
\usepackage[citecolor=red,urlcolor=red]{hyperref}
\hypersetup{
	colorlinks=true,
	linkcolor=blue,
	filecolor=red,      
	urlcolor=blue,
	citecolor=blue,
}
\usepackage{graphicx} 
\usepackage{subfigure}
\usepackage{colortbl}
\usepackage{hyperref}
\usepackage{nomencl}
\makenomenclature

\def\BibTeX{{\rm B\kern-.05em{\sc i\kern-.025em b}\kern-.08em
		T\kern-.1667em\lower.7ex\hbox{i}\kern-.125emX}}

\usepackage{booktabs}

\usepackage{bm}

\usepackage{algorithm}
\usepackage{algpseudocode}
\usepackage{amsmath}

\begin{document}
	\title{ \textcolor{black} {
	 	 	Rotor-Failure-Aware  Quadrotors Flight \\
	 in Unknown Environments 
}}

	\author{
	\IEEEauthorblockN{Xiaobin Zhou\IEEEauthorrefmark{1\dag}}, 
	\IEEEauthorblockN{Miao Wang\IEEEauthorrefmark{1\dag}}, 
	\IEEEauthorblockN{Chengao Li},
	\IEEEauthorblockN{Can Cui},
	\IEEEauthorblockN{Ruibin Zhang},
	\IEEEauthorblockN{Yongchao Wang},
	\IEEEauthorblockN{Chao Xu},
	\IEEEauthorblockN{Fei Gao\IEEEauthorrefmark{3*}}

	\thanks{*Equal contribution.}
	\thanks{\dag Corresponding author: Fei Gao}

	\thanks{  
		This work was supported in part by the National Natural Science Foundation of China under Grants 62303412 and 62403419; in part by the State Key Laboratory of Advanced Rail Autonomous Operation, Beijing Jiaotong University, under Project RAO2026K09; and in part by the Natural Science Foundation of Zhejiang Province (Huzhou City) under Grant 2023YZ01.

		 Xiaobin Zhou and Miao Wang are with the School of Robotics and Automation, Nanjing University, Suzhou, 215163, China (e-mail: xbzhou\_zju@zju.edu.cn; wmubuntu1991@gmail.com). 	 Can Cui, Ruibin Zhang, Chao Xu, and Fei Gao  are with the Institute of Cyber-Systems and Control, College of Control Science and Engineering, Zhejiang University, Hangzhou 310027, China (e-mail: cuican1990@zju.edu.cn;   ruibin\_zhang@zju.edu.cn;  cxu@zju.edu.cn;  fgaoaa@zju.edu.cn ).  		Chengao Li is with the Department of Aeronautical and Aviation Engineering, The Hong Kong Polytechnic University, Hong Kong, China (e-mail: cheng-ao.li@connect.polyu.hk).  		Yongchao Wang is with the School of Aeronautic Science and Engineering, Beihang University, Beijing, 100191, China (e-mail: wangyongchao@buaa.edu.cn). 
		
		}

}

	\markboth{	IEEE Transactions on Aerospace and Electronic Systems}%
	{Shell \MakeLowercase{\textit{et al.}}:   }
	\maketitle

\begin{abstract}
Rotor failures in quadrotors may result in high-speed rotation and vibration due to rotor imbalance, which introduces significant challenges for autonomous flight in unknown environments. The mainstream approaches against rotor failures rely on fault-tolerant control (FTC) and predefined trajectory tracking. To the best of our knowledge, online failure detection and diagnosis (FDD), trajectory planning, and FTC of the post-failure quadrotors in unknown and complex environments have not yet been achieved. This paper presents a rotor-failure-aware quadrotor navigation system designed to mitigate the impacts of rotor imbalance. 
\textcolor{black}{First, a composite FDD-based nonlinear model predictive controller (NMPC), incorporating motor dynamics, is designed to ensure fast failure detection and flight stability.} Second, a rotor-failure-aware planner is designed to leverage FDD results and spatial-temporal joint optimization, while a LiDAR-based quadrotor platform with four anti-torque plates is designed to enable reliable perception under high-speed rotation. Lastly, extensive benchmarks against state-of-the-art methods highlight the superior performance of the proposed approach in addressing rotor failures, including propeller unloading and motor stoppage. The experimental results demonstrate, \emph{for the first time}, that our approach enables autonomous quadrotor flight with rotor failures in challenging environments, including cluttered rooms and unknown forests.
\end{abstract}

\begin{IEEEkeywords}

Autonomous quadrotor flight, Failure detection and diagnosis, Fault-tolerant control, Rotor failures.
\end{IEEEkeywords}

\IEEEpeerreviewmaketitle

\section{Introduction}

\begin{figure}
	\centering
	\includegraphics[width=1.0\linewidth]{"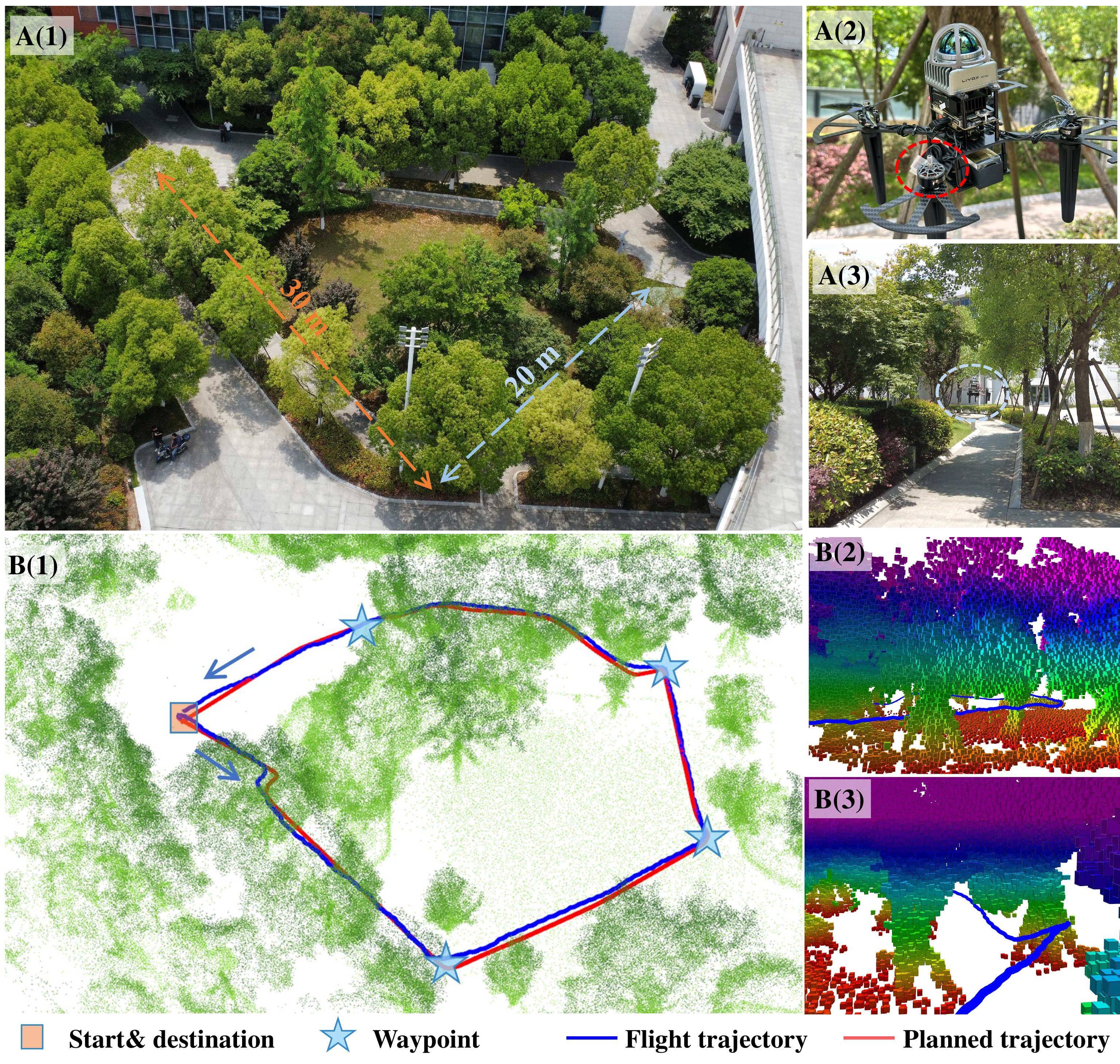"}
	\caption{\textcolor{black}{Autonomous waypoint navigation of a quadrotor with rotor failure in the forest. A(1) is the Bird’s-eye view of the flight environment. A(2) and A(3) show snapshots of third-person view from the real-world experiment, where the quadrotor navigates through narrow passages.  B(1)-B(3) present the 3D point cloud map and detailed views of the experiment. The planned and flight trajectories are depicted as red and blue paths, respectively. The start and destination points are marked with orange squares, while four waypoints are highlighted with light blue stars. The video is available at \url{https://sojustfish.github.io/Rotor-Failure-Aware/}.}}
	\label{outdoor_nav}
\end{figure}

\IEEEPARstart{M}{ultirotors}, renowned for their exceptional maneuverability and flexibility, find applications in a wide variety of scenarios, including field exploration, cave surveys,  and ambitious missions on outer planets \cite{zhou2022swarm, Cui2024FastSim,  tabib2021autonomous}. However, a significant vulnerability of the multirotors is the susceptibility to rotor failures. \textcolor{black}{As reported in \cite{witze2024first}, NASA’s Ingenuity helicopter, featuring a coaxial rotor configuration, suffered rotor blade failures after three years of successful operation, which eventually led to the end of its mission.} Rotor failures not only disrupt normal flight operations but also pose serious safety risks to surrounding humans \cite{ackerman2019swiss}.  To ensure post-failure flight safety, an autonomous and reliable rotor-failure-aware navigation system is essential without peripherals. For the damaged quadrotor to achieve autonomous flight in unknown environments, flight stability, collision avoidance, and dynamic feasibility must be explicitly considered to cope with these practical challenges as follows.

The first challenge lies in swiftly restoring flight stability and achieving precise position tracking following rotor failures. When a quadrotor experiences a sudden rotor failure during normal flight, it faces issues such as attitude control divergence and altitude loss. Both active fault-tolerant control (FTC) \cite{Wu2021Nonlinear, Stephan2018Linear} and passive FTC \cite{Ke2023Uniform } are employed to address rotor failures. Active FTC focuses on real-time failure detection and diagnosis (FDD) and reconfiguring the controller structure to manage faults dynamically. In contrast, passive FTC leverages system redundancies and pre-designed robustness to handle faults without requiring control reconfiguration \cite{zhang2008bibliographical}. Because of the quadrotors' limited redundancies and the conservative nature of passive FTC, active FTC is more extensively studied to improve system performance following rotor failures.

FDD plays a critical role in providing failure information for active FTC. Computation delay in FDD beyond the acceptable threshold can result in the damaged quadrotors, operating with the original controller structure, entering an irreversible state. Therefore, many studies \cite{lippiello2014emergency, Merheb2017Emergency} assume failure information is known a priori and do not account for FDD computation delay. In reality, rotor failures can occur in any of the quadrotors' motors or propellers, making it unobservable without effective FDD mechanisms. Furthermore,  rotor failures, such as propeller damage or motor arm breakage, pose greater challenges for FDD compared to pre-designed, code-trigger motor-stopping failure. Addressing the first challenge requires the development of a carefully designed FDD algorithm alongside the control reconfiguration strategy.

The second challenge lies in ensuring reliable perception and collision avoidance in unknown environments, while accommodating altered dynamic feasibility following rotor failures. To maintain thrust balance, the thrust of the rotor opposite to the failed one is restricted to a lower amplitude \cite{yu2023integrated}. This constraint significantly alters the quadrotor's thrust-to-weight ratio and yaw torque, changing the safe flight envelope and dynamic feasibility \cite{sun2020incremental}. Consequently, trajectory planning must account for the change in dynamic feasibility before and after rotor failures. Additionally, the imbalance in yaw torque induces high-speed rotation and vibration, severely degrading the perception performance of autonomous navigation systems \cite{Sun2021Autonomous}. As a result, many related studies rely on global positioning systems (GPS) or motion capture systems (MCS) for post-failure trajectory tracking \cite{Chenxu, yu2023integrated}, limiting their applicability in unknown and complex environments. In scenarios lacking global information and featuring unknown obstacles, global trajectory tracking without real-time perception cannot ensure flight safety. Addressing the second challenge requires a reliable perception and trajectory planning system that accounts for high-speed rotation and altered dynamic feasibility.

Based on the above observations and analysis, we propose a rotor-failure-aware navigation system for quadrotors in challenging environments. \emph{Firstly}, to enhance the FDD speed for real-world rotor failures during different flight phases, we propose a composite FDD mechanism that integrates both model-based and model-free approaches, by taking onboard revolutions per minute (RPM) and state estimation as inputs.  \textit{Secondly}, to ensure collision avoidance and maintain dynamic feasibility before and after rotor failures, we propose a rotor-failure-aware trajectory planning method based on unconstrained nonlinear programming. We design a LiDAR-based quadrotor platform equipped with four anti-torque plates mounted on the landing gears to reduce the rotational speed and enable reliable perception. \textit{Thirdly}, we conduct simulations and experiments in various scenarios (see Fig.~\ref{outdoor_nav}). The primary contributions of this paper are as follows:

\begin{enumerate}

	\item \textcolor{black}{An FTC system, integrating both composite FDD and nonlinear model predictive controller (NMPC), is proposed to ensure timely response and precise trajectory tracking. }  The proposed method is proven to prevent loss of flight control despite various kinds of physical rotor failures.
	
	\item A rotor-failure-aware planning method is developed to address the collision avoidance problem, which can automatically generate feasible trajectories.  Besides, the reliable quadrotor perception under high-speed rotation is considered by a customized LiDAR-based quadrotor, through mechanical design and sensor selection.

	\item The proposed navigation system is extensively tested in comprehensive simulations and real-world experiments. To the best of our knowledge, we are the first to successfully achieve post-failure quadrotor navigation in unknown and complex environments.

\end{enumerate}

The remainder of this paper is organized as follows. Related work is reviewed in Section~\ref{Related Work}. Sections~\ref{Preliminary} and \ref{SYSTEM OVERVIEW} present the quadrotor dynamics and the overall system architecture. The proposed navigation strategy is introduced in Section~\ref{Control System Against Rotor Failure}. Simulation and real-world experimental results are provided in Section~\ref{SIMULATION and Real-World Experiments}. Finally, concluding remarks are presented in Section~\ref{Conclusion}.

\section{Related Work}
\label{Related Work}

In this section, we review prior work on the robust FTC system 
and autonomous flight for quadrotors under rotor failures. We highlight the challenges of fast FDD, dynamic feasibility, and high-speed rotation, which impact the flight safety of post-failure quadrotors in unknown environments.

\subsection{FTC and FDD Against Rotor Failures }

Despite substantial advances in the safety flight control of quadrotors, safety concerns about rotor failures hinder their widespread application.  In the scenario of losing a complete rotor, only three rotors remain operational, causing the quadrotor to spin uncontrollably along the yaw axis and experience high-speed rotation and severe vibration. To address these issues, several control strategies are proposed, including proportional-integral-derivative controller \cite{lippiello2014emergency, Merheb2017Emergency}, linear quadratic regulator \cite{Mueller2014Stability}, and incremental nonlinear dynamic inversion \cite{sun2020incremental}. The main idea of these methods is to maintain altitude and position control by sacrificing yaw control. However, rotor failure information is generally assumed to be known rather than acquired through real-time FDD in the above works. The research on FDD primarily focused on identifying motor and propeller damage, with methods broadly categorized as model-based or model-free approaches  \cite{liu2024audio, o2024learning,mao2024propeller }. 
Unfortunately, the model-free FDD method in \cite{ liu2024audio} is limited to experimental validation on minor propeller damage. Model-based methods aim to quantify motor and propeller damage by leveraging the quadrotor’s model and state estimation. For instance, \cite{mao2024propeller} proposes a method to assess partial propeller damage by analyzing the relationship between motor speed and thrust, coupled with a mismatch index. In contrast to a partial rotor failure, FDD and FTC for a complete rotor failure impose stricter requirements on detection efficiency. To address this, \cite{yu2023integrated} introduces a fixed-time observer to estimate failures induced by code-injected motor stoppage, while \cite{Ke2023Uniform} presents a uniform passive FTC for the quadrotor with up to three rotor failures without relying on fault information. Despite these advances, existing works struggle with addressing the fast detection of both propeller and motor failures across different flight stages. Most current approaches either specialize in detecting a single type of failure or are limited to specific operating conditions, such as code-triggered motor failure during hovering or slow-speed flight, where system dynamics are relatively stable.  Furthermore, the dynamic variability of flight stages ranging from takeoff to trajectory tracking adds complexity to the design of unified fault detection systems.  

\subsection{Post-Failure Flight in Unknown Environments }
Autonomous navigation flight under rotor failures is a challenge for quadrotors in complex environments, due to the degraded dynamic feasibility and high-speed body rotation. Early studies in this field primarily focus on trajectory planning under partial actuator failures (e.g. blade tip loss, motor efficiency loss) \cite{nam2024fault, zhou2024internal, Liu2023Simultaneous,  Flatness2012Chamseddine} and external disturbances (e.g. suspended payload,  air drag) \cite{Li2022AutoTrans, wang2022neither, Wu2021External, Shi2022Neural}. Recent advancements in trajectory planning incorporate maximum thrust limits to analyze maneuverability change \cite{zhou2024internal}. Nonetheless, they rely on a simplified dynamic model and MCS, limiting the practicality in unknown environments. Furthermore, rotor failures enable the quadrotors to rotate at a high speed, which causes motion blur in cameras and skewed point cloud data in LiDAR-based perception systems. To address this challenge, Sun et al. \cite{Sun2021Autonomous} compare the effectiveness of event cameras and standard cameras for state estimation in post-failure quadrotors. The experimental results demonstrate that the event camera provides reliable localization under low-light conditions and enables trajectory tracking in outdoor environments. Besides, Chen et al. \cite{chen2023self} develop a self-rotating aerial robot with increased perception capability and endurance, which achieves autonomous navigation with high-speed body rotation. Yet, the trajectory planning approach in \cite{chen2023self} does not involve the system dynamics change from rotor failures. In summary, autonomous navigation in unknown environments for rotor-failure quadrotors remains a challenging problem, due to the degraded dynamic feasibility and high-speed rotation. Furthermore, the integration of FDD results into autonomous perception, planning, and control has yet to be demonstrated on a post-failure quadrotor.

\section{ Preliminaries }
\label{Preliminary}
\subsection{ Notations }

\begin{figure}
	\centering
	\includegraphics[width=0.90\linewidth]{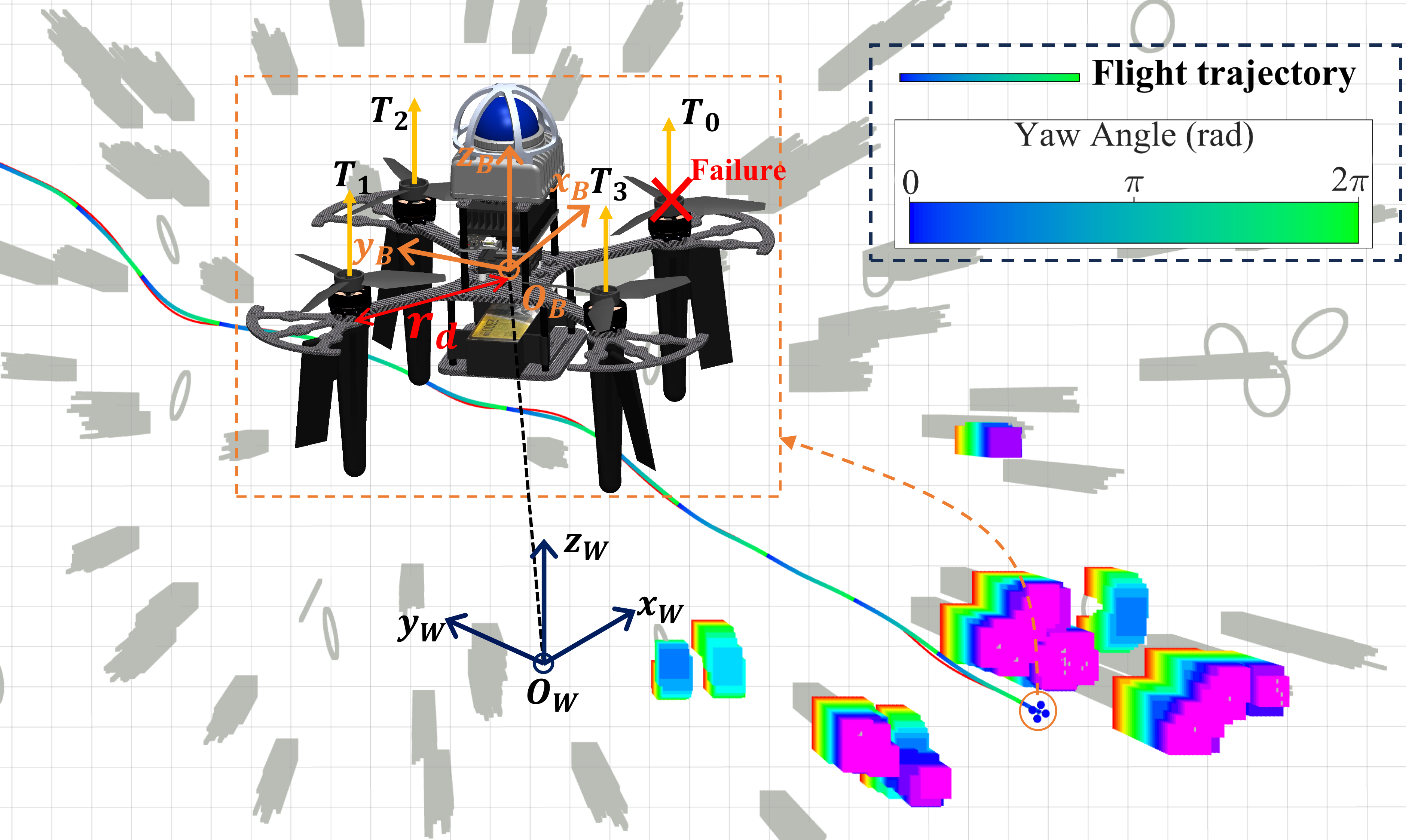}
	\caption{\textcolor{black}{The illustration of thrust index, inertial coordinates, body coordinates, and flight trajectory for the quadrotor with rotor failures.}}
	\label{quadrotor_model}
\end{figure}

Throughout this paper, the set of real numbers is denoted as \(\mathbb{R}\), while \(\text{diag}[d_1, d_2, d_3]\) refers to a diagonal matrix with \(d_1\), \(d_2\), and \(d_3\) on its diagonal.   For a matrix \(\bm{A} \in \mathbb{R}^{m \times n}\) or a vector \(\bm{x} \in \mathbb{R}^n\), the transpose is denoted by \(\bm{A}^T\) or \(\bm{x}^T\), respectively. The Euclidean norm of \(\bm{x}\) is written as \(\|\bm{x}\|\), defined by \(\|\bm{x}\| = \sqrt{\bm{x}^T \bm{x}}\). 
The coordinate systems of the quadrotor consist of two right-handed frames: the inertial frame, \( F_w: \{\bm{x}_w, \bm{y}_w, \bm{z}_w\} \), with \(\bm{z}_w\) pointing upward, opposite to gravity; and the body frame, \( F_B: \{\bm{x}_B, \bm{y}_B, \bm{z}_B\} \), where \(\bm{x}_B\) points forward and \(\bm{z}_B\) aligns with the thrust direction (see Fig. \ref{quadrotor_model}).  The matrix and vector with the subscript \( B \) are represented in the body frame, while the matrix and vector without a subscript are represented in the inertial frame. The transformation from \( F_w \) to \( F_B \) is defined by the rotation matrix \( \bm{R} \), which is parameterized by the quaternion \( \bm{q} = [q_w, q_x, q_y, q_z]^T  \).

\begingroup
\linespread{1.5}
\begin{table}[!t]
	\centering
	\caption{Symbols Used in this Paper.}
	\label{tab_symbols}
	\begin{tabular}{@{}ll@{}}
		$ F_B, F_w$                             & Quadrotor frame and world frame. \\
		$\bm{x}_w, \bm{y}_w, \bm{z}_w \in \mathbb{R}^3$          & Unit vectors  of the world frame. \\
		$\bm{x}_B, \bm{y}_B, \bm{z}_B \in \mathbb{R}^3 $        & Unit vectors  of the quadrotor frame. \\
		$\bm{\eta}, \bm{v} \in \mathbb{R}^3$                      & Position and linear velocity of the quadrotor.            \\
		$\bm{w}_B = [p, q, r]^T \in \mathbb{R}^3 $               & The angular velocity of the quadrotor.            \\
		$\bm{\alpha}_{B} \in \mathbb{R}^3$                       & The angular acceleration of the quadrotor.  \\
		$\bm{R} \in SO(3)$                                & Rotation matrix of the quadrotor.      \\
			$\bm{D} \in \mathbb{R}^{3\times3}$                                &The rotor-drag coefficients.      \\
		$\bm{\tau} \in \mathbb{R}^3, T\in \mathbb{R}$                          & Torque and total thrust of the quadrotor.\\
		$m_{0} \in \mathbb{R}, \bm{g} \in \mathbb{R}^3  $                           & Mass of the quadrotor and gravity vector.                    \\
		$\bm{t} \in \mathbb{R}^4$                                &  Thrust vector of the quadrotor.    \\
		$\bm{I}_v \in \mathbb{R}^{3\times3}$                              & Inertia matrix of the quadrotor. \\
		$ k_{d, \psi}\in \mathbb{R}$                          & The aerodynamic yaw damping coefficient.  \\
		$k_{n} \in \mathbb{R} $                         & The thrust coefficient.  \\
			$ k_{t} \in \mathbb{R} $                    & The  yaw torque coefficient.  \\
		$w_i\in \mathbb{R}$                                   & The rotation speed of the  motor.\\
		$u_i\in \mathbb{R} $                                  & The command input of the rotor.         \\
	\end{tabular}
\end{table}
\endgroup

\subsection{Quadrotor Dynamics }

The quadrotor is modeled based on the rigid body kinematic and dynamic equations as follows:
\begin{equation}
m_{0}\ddot{\boldsymbol{\eta}} = T \bm{z}_B - \bm{RD}\bm{R}^{T}\bm{v} + m_{0} \bm{g},
\label{kinematic_model}
\end{equation}

\begin{equation}
\dot{\bm{q}} = \frac{1}{2} \bm{q} \otimes \begin{bmatrix} 0 \\ \bm{w}_B \end{bmatrix},
\label{dyna_model}
\end{equation}

\begin{equation}
\bm{I}_v \bm{\alpha}_B =  \bm{\tau} -\bm{w}_B \times \bm{I}_v \bm{w}_B - \bm{A}(r),
\label{torque_model}
\end{equation}
where \( \bm{\eta} \) represents the position of the quadrotor's center of gravity (CoG). $\bm{v}$, \( T \), $\bm{D}$, \( m_{0} \), and \( \bm{g} = [0, 0, -g_{0}]^T \)  denote the velocity vector, total thrust, rotor-drag coefficient, the mass, and  the gravity vector. In the rotational dynamics, \( \otimes \) denotes quaternion multiplication, \( \bm{w}_B = [p, q, r]^T \) represents the angular velocity of frame \( \mathcal{F}_B \) relative to \( \mathcal{F}_w \), and \( \bm{\alpha}_{B} \) is the corresponding angular acceleration. The term \( \bm{I}_v \) denotes the quadrotor's inertia matrix, while \( \bm{\tau} = [\tau_{x}, \tau_{y}, \tau_{z}]^T\) is the total torque generated by the rotors. Additionally, \( \bm{A}(r)= [0, \ 0, \ k_{d, \psi} r]^T \) represents the torque induced by rotation motion, and \( k_{d, \psi} \) is the aerodynamic yaw damping coefficient.
 Details of the variables and parameters are listed in the  Table \ref{tab_symbols}. \textcolor{black}{It is noted that the above dynamic model does not capture certain unmodeled effects such as aerodynamic uncertainties, wind disturbance, and sensor noise. While these factors may affect state estimation and control accuracy, the proposed framework mitigates them through the integration of LiDAR-inertial odometry and NMPC. }

The total thrust and rotor-induced torques depend on the thrusts from the four rotors, expressed as
\begin{equation}
\begin{bmatrix}
T \\ \bm{\tau}
\end{bmatrix}
=
\bm{M}_t \bm{t},
\end{equation}
where \( \bm{t} = [T_0, \ T_1, \ T_2, \ T_3]^T \) represents the thrust vector,  with each rotor thrust defined as  $ 0 \leq  T_i = k_{n} w_i^2  \leq  \overline{T}$, $\overline{T}$ is the thrust limit of each motor, and \( i = 0, 1, 2, 3 \). Here, \( k_{n} \) is the  thrust coefficient of the propeller, and \( w_i \) is the rotation speed of the \( i \)-th motor. \textcolor{black}{The control effectiveness matrix \( \bm{M}_t \) is given by:
\begin{equation}
\bm{M}_t = \begin{bmatrix}
1 & 1 & 1 & 1 \\
- \frac{\sqrt{2}}{2} r_{d} & \frac{\sqrt{2}}{2} r_{d} & \frac{\sqrt{2}}{2} r_{d} & -\frac{\sqrt{2}}{2} r_{d} \\
-\frac{\sqrt{2}}{2} r_{d} & \frac{\sqrt{2}}{2} r_{d} & -\frac{\sqrt{2}}{2} r_{d} & \frac{\sqrt{2}}{2} r_{d} \\
-\kappa_t & -\kappa_t & \kappa_t & \kappa_t
\end{bmatrix},
\end{equation}
where $r_{d}$ is the half roll or pitch rotor-to-rotor distance,  and \( \kappa_t \) denotes the yaw torque coefficient.} For real-time control applications, the motors exhibit a delay in reaching commanded thrust levels. To account for this, we incorporate motor dynamics into the model as
\begin{equation}
\dot{T}_i = \frac{1}{\sigma}(u_i - T_i),
\end{equation}
where \( \sigma \) is a time constant derived from empirical data and \( u_i \) represents the command input.

\section{SYSTEM OVERVIEW}
\label{SYSTEM OVERVIEW}

\subsection{Software Framework}

\begin{figure*}    
	\centering
	\includegraphics[width=0.80\linewidth]{"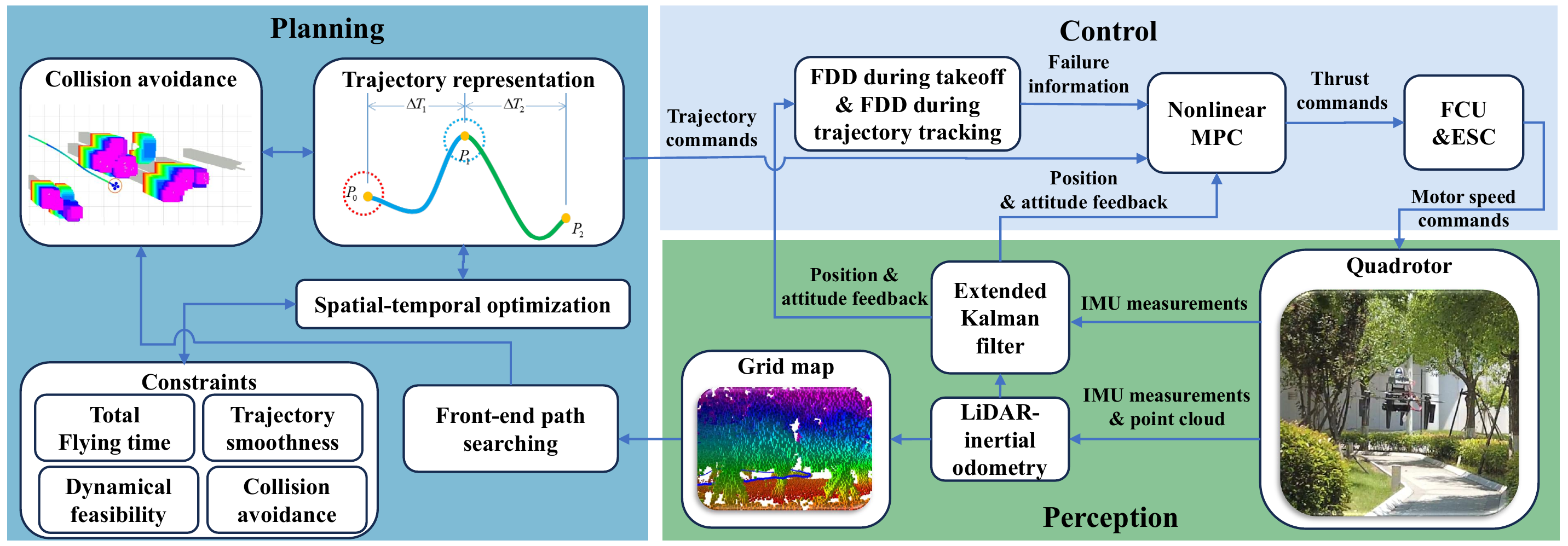"}
	\caption{\textcolor{black}{ The proposed system consists of three main components: perception, planning, and control. 
	}}
	\label{flowchart}
\end{figure*}

To enable autonomous quadrotors flight under rotor failures, the software framework illustrated in Fig. \ref{flowchart}, comprising perception, planning, and control modules, is integrated into the onboard computer. \textcolor{black}{The onboard computation is carried out on an NVIDIA Jetson Orin NX, a powerful embedded and edge computing platform equipped with an 8-core CPU, a 1024-core GPU, and 16 GB of RAM. This hardware runs Ubuntu and ROS, and provides sufficient computational power for real-time NMPC and perception modules.}  Communication between the onboard computer and the flight controller is facilitated via MAVROS \cite{Mavros2024git}. The quadrotor's software system primarily consists of the following subsystems.

\begin{figure}
	\centering
	\includegraphics[width=0.950\linewidth]{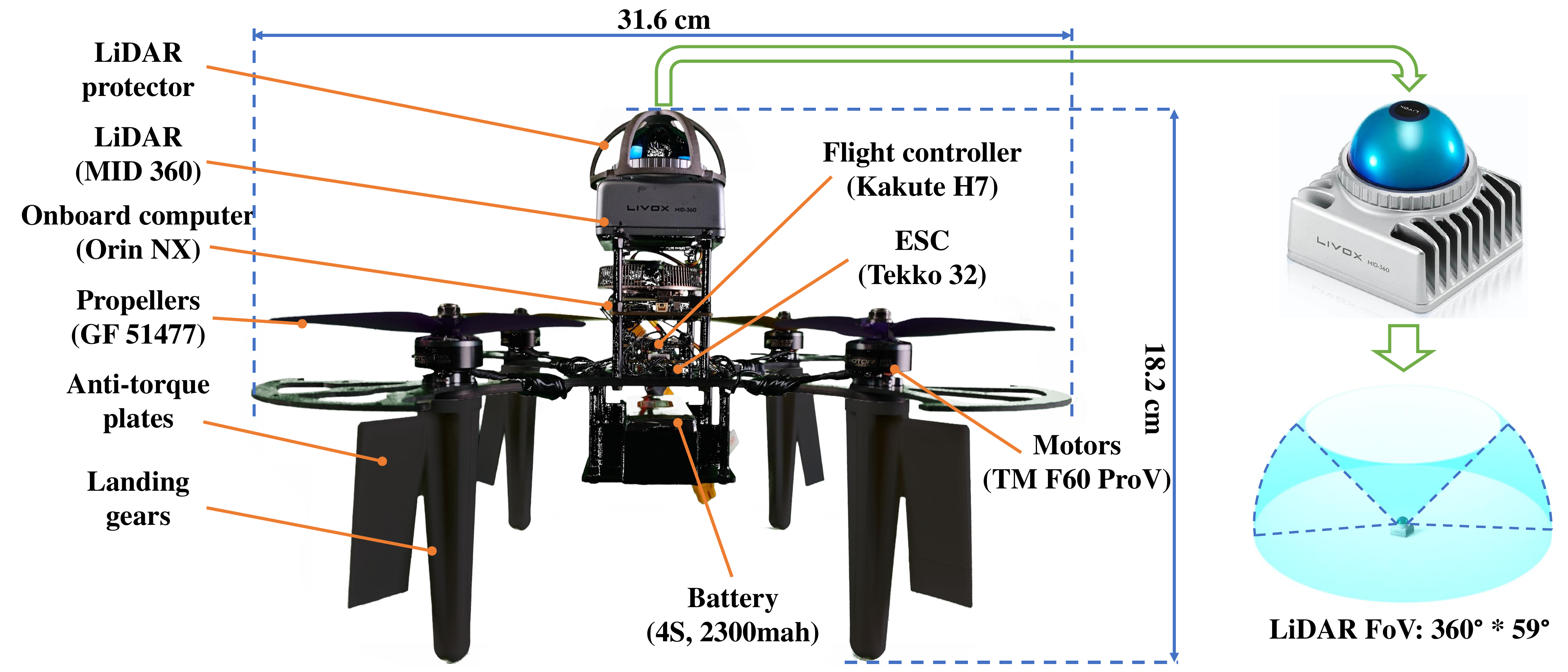}
	\caption{\textcolor{black}{The platform  and components description of the designed quadrotor.}}
	\label{quadrotor_hardware}
\end{figure}

1) Control Unit. The high-level system comprises the FTC framework, including a controller and an FDD module, both operating at 200 Hz on the  NVIDIA Jetson Orin NX and ultimately outputting the desired RPMs for the four motors. The low-level control unit maintains closed-loop control of RPM, which consists of the flight controller (Kakute H7 V1.3) and an electronic speed controller (ESC, Tekko 32 F4). The control process involves the following steps. The flight controller receives the desired RPM commands from the  NVIDIA Jetson Orin NX and transmits digital signals to the ESC via the DShot protocol.
The ESC decodes the DShot signals and adjusts motor speed by controlling the voltage and current. It also measures the actual motor speed using built-in sensors and sends this feedback to the flight controller via the DShot protocol. Based on the feedback RPM data, the flight controller refines the control signals to achieve precise closed-loop motor speed regulation.

2) Planning Unit.  To generate rotor-failure-aware trajectories in complex environments, we propose a planning module integrated into the onboard computer that processes point clouds captured by onboard sensors.  The real-time planner is designed to produce smooth, obstacle-free trajectories based on real-time perception, which is presented in the following subsection.

\newtheorem{remark}{Remark}

\subsection{Quadrotor Platform Design} 

All experiments are performed on a custom-designed and assembled quadrotor platform with a wheelbase of 250 $mm$. The hardware specifications of this platform are presented in Fig. \ref{quadrotor_hardware}. The platform weighs 1150 $g$ and measures 31.6 × 31.6 × 18.2 $cm$. It is powered by a 2300 $mAh$ lithium-polymer battery and equipped with four 2550 $KV$ brushless motors and 5-inch three-blade propellers.  When a rotor failure occurs, the imbalance in yaw torque causes the quadrotor to enter a high-speed spinning motion accompanied by intense vibrations. Under these conditions, the efficiency of state estimation and mapping significantly deteriorates. First, the IMU is subjected to substantial centrifugal acceleration, resulting in increased noise and errors in the measurement of acceleration and angular velocity \cite{Sun2021Autonomous}. Second, the odometry accuracy degrades due to motion blur and scan distortions caused by high-speed motion \cite{chen2023self}. The FAST-LIO2 algorithm \cite{xu2022fast}, an efficient and robust LiDAR-inertial odometry framework, is utilized. Besides, the mechanical design  and sensor selection  contribute to meeting the demands of localization and mapping during the high-speed self-rotation of the post-failure quadrotors.

The spin motion of the quadrotor following rotor failures originates from the counter-torque produced by the rotating propellers, with the yaw torque magnitude expressed as: 
\begin{equation} 
\tau_{z} = \left[-k_t, \ -k_t, \ k_t, \ k_t\right] \left[ T_0, \ T_1, \ T_2, \ T_3\right]^{T}, 
\end{equation}
where $\tau_{z}$ denotes the yaw torque acting on the quadrotor. In the event of rotor failures, the imbalance in $\tau_{z}$ leads to an increase in the spin speed $r$ until the air resistance generated by rotation, $-k_{d, \psi} r$, balances $\tau_{z}$. To mitigate excessive spin speed, a viable strategy is to adjust the torque coefficient $k_t$ and the air resistance coefficient $k_{d, \psi}$. Therefore, we improve the quadrotor platform with four anti-torque plates mounted on the landing gears. These plates increase air resistance and compensate for the motor's counter-torque.  The anti-torque plates have a width of 35 $mm$ and a length of 77 $mm$.  The plates have a large projected area in the vertical plane coinciding with the arm, which generates large air resistance during quadrotor spinning, thereby increasing the air resistance coefficient $k_{d, \psi}$. Moreover, the airfoil design allows for creating torque in the opposite direction of the propeller's counter-torque, thereby reducing the torque coefficient $k_t$.  	\textcolor{black}{ The torque coefficient $k_{t}$ is obtained from a power test using a professional test bench, provided by Shenzhen Deep Exploration Technology Co., Ltd (http://www.szde-tech.com/). The yaw damping coefficient is identified from rotor-failure flight experiments. These procedures ensure that both parameters reflect the actual quadrotor dynamics. }
The Livox MID-360 LiDAR and an IMU are selected as the primary sensors (see Fig. \ref{quadrotor_hardware}).  The perception under high-speed rotation is alleviated through the 360° horizontal field-of-view (FoV)  of the Livox MID-360 LiDAR for omni-directional environmental perception. Simultaneously, the IMU operates at a frequency of 333 Hz, providing estimates of the quadrotor's acceleration and angular velocity. Data from both the LiDAR and IMU are processed by an onboard extended Kalman filter (EKF).

\section{	Rotor-Failure-Aware Flight System Design}
\label{Control System Against Rotor Failure}
In this section, we propose a rotor-failure-aware flight system, enabling safe and stable flight of the post-failure quadrotors. We introduce an NMPC-based controller to compute the control commands by solving an optimal control problem in a receding horizon manner (Section \ref{NMPC design}). We cannot ignore the influence of FDD delay on control reconfiguration. Therefore, we propose a composite FDD with onboard data of RPM and LiDAR odometry (Section \ref{Composite FDD for different flight Stages}). Lastly, we present a polynomial trajectory representation method and formulate a nonlinear trajectory optimization problem that incorporates multiple constraints (Section \ref{Rotor-Failure-Aware Trajectory Planning}).

\subsection{NMPC Design}
\label{NMPC design}

By defining the quadrotor state as $\bm{x} = [\bm{\eta}^{T}, \bm{q}^{T}, \dot{\bm{\eta}}^{T}, \bm{w}^{T}, \bm{t}^{T}]^{T}$ and the control input as $\bm{u} = [ u_{0}, u_{1}, u_{2}, u_{3}]^{T}$, the quadrotor dynamics can be expressed in discrete-time form as $\bm{x}_{k+1} = f(\bm{x}_{k}, \bm{u}_{k})$. The NMPC framework seeks to minimize a cost function defined by the errors between the reference and predicted states over a finite-time horizon $t_N \in [t, t+h]$. The step size is determined by $dt = h/N$, where $N$ represents the number of time steps and $h$ is the horizon length. The  constrained optimization problem can be formulated as:
\begin{equation}
	\begin{aligned}
		&{\bm{u}_{k:k+N-1}} = \min_{\bm{u}} \bm{y}_N^T \bm{Q}_N \bm{y}_N + \sum_{i=k}^{k+N-1} \bm{y}_i^T \bm{Q} \bm{y}_i + \bm{u}_i^T \bm{R}_N \bm{u}_i \\
		& \text{s.t.} \quad \bm{x}_{i+1} = f(\bm{x}_i, \bm{u}_i), \quad i = k, k+1, \ldots, k+N-1 \\
		& \quad \quad \underline{\bm{u}} \leq \bm{u} \leq \overline{\bm{u}}, \quad  \underline{\bm{w}} \leq \bm{w} \leq \overline{\bm{w}}.
	\end{aligned}
\end{equation}
Here, $k$ denotes the current time step,  and $\bm{x}_{k:k+N}$ refers to the predicted state trajectory. The weighting matrices $\bm{Q}$, $\bm{Q}_{N}$, and $\bm{R}_N$ are used to tune the optimization process. $ \underline{\bm{u}}$, $\overline{\bm{u}}$, $ \underline{\bm{w}}$, and $\overline{\bm{w}}$ represent the lower and upper bounds of the control input and angular velocity, respectively.

Under normal flight conditions, the control input limit is set as $\overline{\bm{u}} = T_{\text{max}} \bm{1}_{4 \times 1}$. If a rotor failure is identified by the fault detection module, the $i$-th element of $\overline{\bm{u}}$ is updated to reflect the fault condition by setting $\overline{u}_i = u_i = 0$ for the failed rotor.

For a quadrotor experiencing rotor failures, achieving full attitude control becomes unfeasible, necessitating the sacrifice of controllability in the yaw channel. \textcolor{black}{ To address this, a tilt-prioritized control strategy proposed in \cite{Brescianini2020Tilt} is employed to manage the reduced-attitude error (i.e., pitch and roll channels) $\tilde{\bm{q}}_{xy}$ and the yaw error $\tilde{\bm{q}}_{z}$ independently. The attitude error is defined as:
\begin{equation}
	\tilde{\bm{q}} = \bm{q}_r \otimes \bm{q}^{-1} = [q_{e,w}, \ q_{e,x}, \ q_{e,y}, \ q_{e,z}]^T,
\end{equation}
where $\bm{q}_r$ represents the reference attitude, and $\bm{q}$ denotes the current attitude.} Next, the attitude error is decomposed into its $z$ and $xy$ components, $\tilde{\bm{q}} = \tilde{\bm{q}}_z \circ \tilde{\bm{q}}_{xy}$.

		Using the reduced-attitude error vector $\tilde{\bm{q}}_e = [\tilde{q}_{xy,x}, \ \tilde{q}_{xy,y}, \ {q}_{e,x}^2 + {q}_{e,y}^2, \ \tilde{q}_{z,z}]^T$, the cost vector $\bm{y}_i$, the block-diagonal weight matrices $\bm{Q}$, and $\bm{Q}_N$ for the running costs are constructed as follows:

		\begin{equation}
			\label{eq_5}
			\bm{y}_{i} = \left[ \begin{matrix} 
				\bm{\eta} - \bm{\eta}_{ref} ,
				\tilde{\bm{q}}_e ,
				\bm{v} - \bm{v}_{ref} ,
				\bm{w} - \bm{w}_{ref} ,
				\bm{t} - \bm{t}_{ref} 
			\end{matrix}
			\right]^T,
		\end{equation}
		
		\textcolor{black}{
		\begin{equation}
			\label{eq_5}
			\bm{Q} = \text{diag} ( [\bm{Q}_{p}, \bm{Q}_{q}, \bm{Q}_{v}, \bm{Q}_{w}, \bm{Q}_{t} ]),
		\end{equation} }
		
		\begin{equation}
			\label{eq_5}
			\bm{Q}_{N} = \bm{Q}.
		\end{equation}

\subsection{ Composite FDD for Different Flight Stages}
\label{Composite FDD for different flight Stages}

Rotor failures in quadrotors primarily occur in two critical components:  motors and propellers. Motor failure typically results from overheating caused by excessive load or inadequate cooling, while propeller failure is often due to physical damage from impacts or material fatigue. To address rotor failures, active FTC methods rely on the rapid detection of such failures as a prerequisite. This paper introduces a composite  FDD system that integrates model-based and model-free approaches to enhance both diagnosis speed and accuracy.

\textcolor{black}{ 
For motor failure diagnosis, we employ the real-time motor speed $\hat{\omega}_i$ reported by the ESC via DShot telemetry. 
The estimated speed is compared against the reference motor speed $\omega_i$, and the motor failure index is defined as:
\begin{equation}
M_{i} =\hat{w}_i / {w}_i.
\label{eq:terrestrial_dynamics}
\end{equation}
When the condition $M_i \leq \gamma_M$ is satisfied, where $0 < \gamma_M \leq 1$ denotes the motor failure threshold, the control mode is switched to FTC mode.
}

For diagnosing propeller failure during trajectory tracking,  observers for translational and rotational dynamics are designed as follows:
\begin{equation}
\hat{ \bm{T}}_a  =   T_{f} \bm{z}_B + m_{0} \bm{g} -m_{0} \ddot{ \bm{\eta}}_{f}  - \bm{RD}\bm{R}^{T}\bm{v} ,
\end{equation}
\begin{equation}
\hat{\bm{\tau}}_{a} =  \bm{\tau}_{f} - I_v \bm{\alpha}_{B, f} - \bm{w}_{B, f} \times I_v \bm{w}_{B, f}  - \bm{A}(r),
\end{equation}
where $\hat{ \bm{T}}_a$ and $\hat{\bm{\tau}}_{a}$ represent the estimated loss of thrust and torque, respectively. Here, $T_{f}$ and $ \bm{\tau}_{f}$ denote the estimated thrust and torques, with the thrust of each rotor obtained via rotor speed measurement. Variables $\ddot{ \bm{\eta}}_{f}$, $\bm{\alpha}_{B, f}$, and $\bm{w}_{B, f}$ are the measured acceleration, angular acceleration, and angular velocity from the low-pass filter. The combined thrust and rotor-induced torques depend on the thrusts of the four rotors as:
\begin{equation}
\begin{bmatrix}
||\hat{ \bm{T}}_a || \\ \hat{\bm{\tau}}_{a}
\end{bmatrix}
=
\bm{M}_t \bm{t}_{*},
\end{equation}
where $\bm{t}_{*} = [\hat{T}_0, \ \hat{T}_1, \ \hat{T}_2, \ \hat{T}_3]^T$ represents the thrust loss for the four rotors, and $\bm{M}_t$ is the control effectiveness matrix. Propeller damage is quantified using the following index:
\begin{equation}
P_{i} = \hat{T}_i / \overline{T}.
\label{eq:terrestrial_dynamics}
\end{equation}
When $P_{i}  > \gamma_{P}$ where $0 < \gamma_{P} \leq1 $ serves as the trigger threshold, the $i$th propeller is classified as failed and isolated. The failure information of the $i$th propeller is subsequently transmitted to trigger controller reconfiguration and trajectory planning.

During takeoff, the quadrotor typically operates at a low altitude, leaving minimal room for recovery or corrective maneuvers in the event of a rotor failure. Compared to the trajectory tracking phase, the available reaction time during takeoff is significantly reduced, increasing the likelihood of a crash. To address this, a low-pass filter-based failure detection method is designed, utilizing acceleration $\ddot{\bm{\eta}}_{f} = [\ddot{\eta}_{f,x}, \ \ddot{\eta}_{f,y}]^T$ and angular acceleration $\bm{\alpha}_{Bf} = [{\alpha}_{Bf,x}, \ {\alpha}_{Bf,y}]^T$ for timely failure detection.

For an ideal autonomous vertical takeoff without rotor failures, the quadrotor rapidly increases motor speed to generate sufficient vertical acceleration to reach the desired altitude while ensuring equal thrust among four motors to avoid lateral movement. However, in the event of both motor and propeller failures during takeoff, the quadrotor experiences a pronounced state response abnormality, including significant angular and linear accelerations. According to the above observations, the failure of the Rotor \#$0$ can be detected using the following index:
\begin{equation}
Q_{0} =
\begin{cases}
1, & \text{if } \ \ddot{\eta}_{f,x}  \geq \gamma_{Q, 0},  \\ 
& \quad \ddot{\eta}_{f,y}  \leq -\gamma_{Q, 1},\\
0, & \text{otherwise},  
\end{cases}
\end{equation}
where $\gamma_{Q, 0},\gamma_{Q, 1}$ are the trigger thresholds. The angular accelerations ${\alpha}_{Bf,x}$ and ${\alpha}_{Bf,y} $ can be also used to compared with trigger thresholds $\gamma_{Q, 2},\gamma_{Q, 3}$ to determine the rotor failure.  When $Q_{0} = 1$, the failure of Rotor \#$0$  is determined. Similarly, failures in Rotor \#$1$, \#$2$, and \#$3$ can be detected with filtered acceleration  and angular acceleration.

	\textcolor{black}{ Beyond complete rotor failures, the FDD framework can also detect the partial thrust reductions (e.g., due to wear, overheating, or intermittent performance) by combining a motor-side cue and a propeller-side cue. Intermittent degradations are detectable if they persist within a short observation window or appear with sufficient frequency. Certain mechanical anomalies such as shaft misalignment typically manifest as speed–thrust inconsistencies and are therefore also within the detectable scope of the proposed FDD. Extremely subtle or very short-lived anomalies may remain undetected, suggesting future extensions that incorporate vibration or thermal cues. }

\subsection{Rotor-Failure-Aware Trajectory Planning}
\label{Rotor-Failure-Aware Trajectory Planning}

In motion planning, the initial collision-free path is generated using the dynamic A* algorithm \cite{zhou2019robust} as the front-end, while the proposed planning method refines the trajectory in the back-end. For the optimization process, the rotor-failure-aware planning framework is designed to adjust spatio-temporal trajectories efficiently. This framework enables linear-complexity deformation of flat-output trajectories, which can then be used to reconstruct the full state as defined in Eqs. \ref{kinematic_model}-\ref{torque_model}.

For a polynomial trajectory with $M$ segments in $m$ dimensions and a degree $N_{0}=2s-1$ (where $s=3$ is the integrator chain order), a linear-complexity mapping is defined as:
\begin{equation}
\bm{c} = M(\bm{q}^{p}, \bm{T}^{p}),
\label{eq:polynomial_map}
\end{equation}
where $\bm{c} = (\bm{c}_1^T, \ldots, \bm{c}_M^T)^T $ represents the polynomial coefficient matrix \cite{zhou2022swarm}. The mapping $M(\bm{q}^{p}, \bm{T}^{p})$ smoothly relates the intermediate waypoints $\bm{q}^{p} = (\bm{q}_1, \ldots, \bm{q}_{M-1})^T $ and time allocation $\bm{T}^{p} = (T_1, \ldots, T_M)^T $ to $\bm{c}$. Using $M(\bm{q}^{p}, \bm{T}^{p})$, the optimization variables are transformed into $\{\bm{q}^{p}, \bm{T}^{p}\}$. The overall trajectory $\bm{p}(t)$ and its individual segment $\bm{p}_g(t)$ are expressed as:
\begin{equation}
\bm{p}(t) = \bm{p}_g (t - t_{g-1}), \quad \forall t \in [t_{g-1}, t_g],
\end{equation}
\begin{equation}
\bm{p}_g(t) = \bm{c}_g^T \bm{\beta}(t), \quad \forall t \in [0, T_g],
\end{equation}
where $ \bm{\beta}(t) = [1, t, \ldots, t^{N_{0}}]^T$ denotes the natural polynomial basis. The trajectory generation for the quadrotor is formulated as an unconstrained nonlinear optimization problem:
\begin{equation}
\min_{\bm{q}^{p}, \bm{T}^{p}} [J_t, J_s, J_d, J_c] \cdot \bm{\lambda}^{p},
\end{equation}
where $J_t$, $J_s$, $J_d$, and $J_c$ penalize total time, trajectory smoothness, dynamic feasibility, and collision avoidance, respectively. The weighting vector $\bm{\lambda}^{p}=[\lambda_t, \lambda_s, \lambda_d, \lambda_c]^T$ balances the trade-offs among these objectives.

To enhance navigation efficiency, the total time is minimized as:
\begin{equation}
J_t = \sum_{g=0}^M T_g,
\end{equation}
where \(T_g\) represents the duration of the \(g\)-th trajectory segment.

The smoothness penalty \( J_s \) is defined as the integral of the squared \(s\)-th order derivatives of the trajectory:
\begin{equation}
J_s = \int_{t_0}^{t_M} \|p^{(s)}(t)\|_2^2 \, \mathrm{d}t.
\end{equation}
Since the trajectory is represented as a piecewise polynomial, this integral can be efficiently computed analytically.

In reference to \cite{zhou2022swarm}, we impose limits on linear velocity, acceleration, and jerk to ensure dynamic feasibility during trajectory generation. The corresponding cost function is defined as:
\begin{equation}
J_d = J_{d,v} + J_{d,a} + J_{d,j},
\end{equation}

\begin{equation}
J_{d,v} = \sum_{i=1}^k  \max\left\{ { \dot{p}^2(t_i)} - v_{{max}}^2, 0\right\},
\end{equation}

\begin{equation}
J_{d,a} = \sum_{i=1}^k  \max\left\{ { \ddot{p}^2(t_i)} - a_{{max}}^2, 0\right\},
\end{equation}

\begin{equation}
a_{{max}}=
\begin{cases}
a_{ {max}, f},  \text{if failure is detected by FDD}  \\
a_{ {max}, n}, \text{otherwise} 
\end{cases}
\label{eq:acceleration_limit}
\end{equation}

\begin{equation}
J_{d,j} = \sum_{i=1}^k  \max\left\{ { \dddot{p}^2(t_i)} - j_{{max}}^2, 0\right\},
\end{equation}
where \(v_{{max}}\) and \(j_{{max}}\) are the predefined velocity and jerk thresholds in normal flight conditions. \(a_{{max, n}}\) and \(a_{{max, f}}\) represent the acceleration thresholds before and after rotor failures, respectively.

To maintain thrust balance along the vertical axis \(\bm{z}_w\) under conditions of zero vertical acceleration (\(a_z = 0\)) and velocity (\(v_z = 0\)), the equations of motion are given as:
\begin{equation}
\label{eq:vertical_balance}
\begin{aligned}
m_{0} a_{xy} &= T z_{B,xy} - f_{xy}, \\
m_{0} a_{z} &= T z_{B,z} - m_{0}g_{0},
\end{aligned}
\end{equation}
where \(z_{B,xy}\) and \(z_{B,z}\) are the horizontal and vertical components of the unit thrust direction vector \(\bm{z}_B\). \(a_{xy}\) represents horizontal acceleration, and \(f_{xy} \) denotes the rotor drag in the horizontal plane. Given \((z_{B,xy})^2 + (z_{B,z})^2 = 1\), the thrust constraint can be expressed as:
\begin{equation}
T^2 - (m_{0} g_{0})^2 = (f_{xy} + m_{0} a_{xy})^2,
\end{equation}

\begin{equation}
a_{xy} = \frac{\sqrt{T^2 - (m_{0}g_{0})^2} - f_{xy}}{m_{0}}.
\end{equation}

To ensure flight safety, we conservatively assume that the rotor opposite to the faulty one is non-functional, resulting in the thrust loss from two rotors for a post-failure quadrotor. For a post-failure quadrotor, the available thrust is constrained by \(0 \leq T \leq T_{f, {max}} = 2 \overline{T}\), and the aerodynamic drag in the horizontal plane satisfies \(f_{xy} \leq f_{{max}} = \|\bm{RD}\bm{R}^T\| v_{{max}}\). Under these constraints, the acceleration limit is updated as:
\begin{equation}
\label{eq_accelera}
a_{{max}, f} = \gamma_{a,f} \left| \frac{\sqrt{T_{f, {max}}^2 - (m_{0} g_{0})^2} - f_{{max}}}{m_{0}} \right|,
\end{equation}
where \(0 < \gamma_{a,f} \leq 1\) is a safety factor to enhance flight reliability. \textcolor{black}{ To ensure safe operation under rotor failure conditions, dynamic feasibility constraints are explicitly considered. The bounds on velocity and jerk are defined in Eqs.~(26) and (29), respectively. Their threshold values are selected empirically based on the characteristics of the quadrotor platform and extensive experimental observations, following a similar practice to that in \cite{zhou2020ego, ren2025safety}. By contrast, the acceleration constraint updated in Eq.~(33) serves as the primary safety limitation in post-failure scenarios, as it directly reflects the reduced actuation capability after rotor degradation.} Although we do not explicitly model efficiency losses or battery degradation, our trajectory optimization already includes total time minimization, which serves as a proxy for energy consumption. This reflects the fact that shorter flight times typically correspond to reduced power usage.

Obstacles are represented as planes defined by the equation:  
\begin{equation}
(\bm{x}^{p} - \bm{s}^{p})^T \bm{v}^{p} = 0, \quad \bm{x}^{p} \in \mathbb{R}^3,
\end{equation}  
where \(\bm{s}^{p} \in \mathbb{R}^3\) denotes a point on the plane, and \(\bm{v}^{p} \in \mathbb{R}^3\) is the normal vector pointing toward the free space. To quantify the relationship between a given point \(\bm{p}^{p} \in \mathbb{R}^3\) and the obstacle, the distance \(d_o\) to the plane is computed as:  
\begin{equation}
d_o = (\bm{p}^{p} - \mathbf{s}^{p})^T \mathbf{v}^{p}.
\end{equation}  
The cost \(J_c\) is expressed as:  
\begin{equation}
J_c = \sum_{i=0}^{\kappa} \max\left\{\left(C_o - d_o(p(t_i))\right), 0\right\}^3,
\end{equation}  
where $C_o$ denotes obstacle clearance and \(d_o\) depends on the position.

\definecolor{mygray}{gray}{.9}
\begin{table}[!t]
	\renewcommand{\arraystretch}{2}
	\caption{Parameters of the Quadrotor Dynamics}
	\label{table_2}
	\centering
	\begin{tabular}{llll}
		\toprule
		Parameter & Value & Parameter & Value \\
		\midrule
	\rowcolor{mygray}	$m_{0}$ [$kg$] & 1.15 & 	 $\bm{I}_v$ [$gm^{2}$] & diag(6.862, 6.992, 8.650)\\
	 $r_{d}$ [$m$] & 0.063 & $k_{n}$ [$N$] & $1.41e^{-8}$ \\
		\rowcolor{mygray}	$k_{d, \psi}$ [$Nms$] & $0.011$ &  $\bm{D}$ [$kg/s$] & diag(0.48, 0.50, 0.65)  \\

		\bottomrule
	\end{tabular}
\end{table}

\begin{figure*}
	\centering
	\includegraphics[width=0.80\linewidth]{"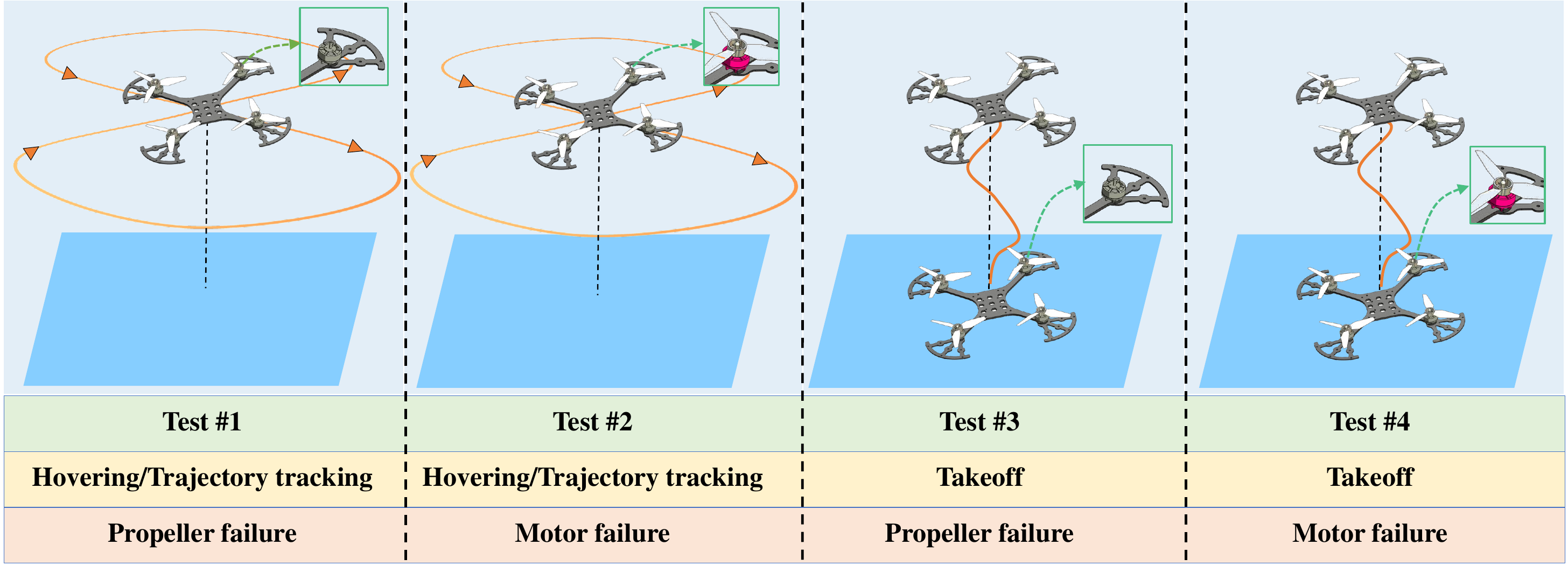"}
	\caption{\textcolor{black}{ Four groups of tests are arranged to verify the robustness of the proposed method under both motor and propeller failure.  In Test \#1 and Test \#2, the rotor failure is injected during trajectory tracking. In Test \#3 and Test \#4, the rotor failure is involved during quadrotor takeoff. }}
	\label{sim_fdd_symbol}
\end{figure*}

\begin{figure}
	\centering
	\includegraphics[width=0.90\linewidth]{"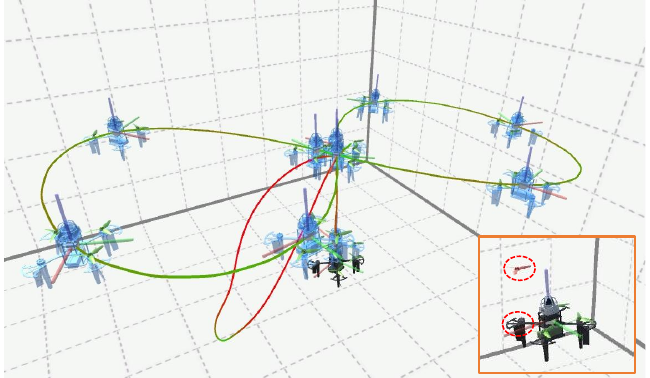"}
	\caption{\textcolor{black}{Simulation visualization \cite{ Cui2024FastSim} with the proposed method in Test \#1, where the propeller-ejection failure occurs during quadrotor hovering. The green and red paths indicate the reference lemniscate trajectory and the real flight trajectory, respectively.}}
	\label{sim_perpeller_ejection}
\end{figure}

\section{  SIMULATION and Real-World Experiments}
\label{SIMULATION and Real-World Experiments}

We evaluate the effectiveness of the proposed method through a combination of simulations and real-world experiments. This section focuses on addressing three critical questions: i) Can the control system handle motor and propeller failures, achieving flight stability without prior knowledge of the failure? (Refer to Sections \ref{Simulation of FDD and FTC } and \ref{FDD and FTC under Different Failure}.) ii) Can the navigation module generate dynamically feasible and collision-free trajectories in cluttered environments, even with unknown obstacles and low-light conditions? (Refer to Section \ref{Autonomous navigation in dynamic and low-light environments}.) iii) How does the rotor-failure-aware framework perform in the wild? (Refer to Section \ref{Autonomous navigation in Dense Garden}.) The physical parameters of the quadrotor are listed in Table~\ref{table_2}. For trajectory planning, the unconstrained optimization problem is solved using LBFGS \cite{liu1989limited}. 
 The NMPC problem is handled by the ACADO \cite{houska2011acado} toolkit in conjunction with qpOASES \cite{ferreau2014qpoases}, operating at a frequency of 200~Hz. The horizon length \(N\) and the time step \(dt\) are set to 20 and 50 $ms$, respectively.  The parameters of the FTC system are specified as  	$ \bm{Q}_{p}= \mathrm{diag}(100, 100, 600)$, $ \bm{Q}_{v} = \mathrm{diag}(5, 5, 5)$, $ \bm{Q}_{q} =\mathrm{diag}(60, 60, 60, 60)$, $ \bm{Q}_{w} = \mathrm{diag}(1, 1, 1)$, $ \bm{Q}_{t} = \mathrm{diag}(1, 1, 1, 1)$, $ \bm{R}_N = \mathrm{diag}(1, 1, 1, 1)$, $ \gamma_{M} = 0.2  $, $\gamma_{P}$ = $0.8$. The weighting vector $\bm{\lambda}^{p}=[10.0, 1.0, 10000.0, 10000.0]^T$ balances the trade-offs among optimization objectives and the safe distance is 0.3 $m$.
 \textcolor{black}{To quantitatively evaluate the tracking performance,  let $\bm{\eta}_t = [x_t, y_t, z_t]^T$ denote the UAV position at time $t$, and $\bm{\eta}_r = [x_r, y_r, z_r]^T$ the corresponding reference position. The root mean square error (RMSE) of trajectory tracking is given by  
\begin{equation}
	e_t = \frac{\sqrt{\int_{t_1}^{t_2} \lVert \bm{\eta}_t - \bm{\eta}_r \rVert^2 \, dt}}{t_2 - t_1},
\end{equation}
where $[t_1, t_2]$ represents the evaluation time interval. }

\begin{figure}
	\centering
	\includegraphics[width=1.0\linewidth]{"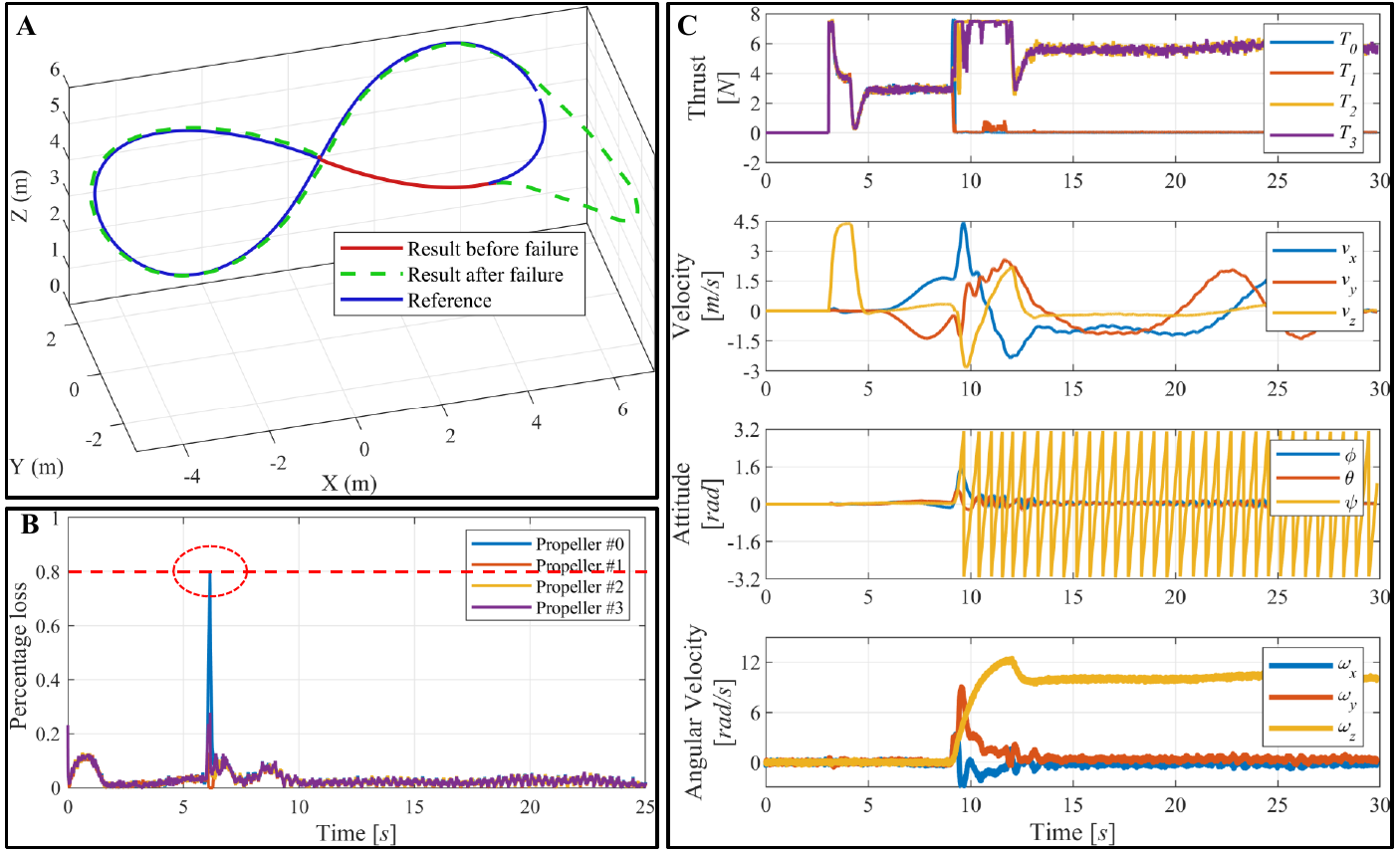"}
	\caption{\textcolor{black}{FDD and FTC results using the proposed method in Test \#1  where the failure of Propeller \#0 occurs during quadrotor trajectory tracking. (A) The visualization of the reference and flight trajectories. (B)  The estimation result of the thrust coefficient loss for four propellers. (C) Time history of thrust and state responses controlled by the proposed FTC.}}
	\label{sim_fdd_propeller}
\end{figure}

\subsection{Simulation of FTC System }
\label{Simulation of FDD and FTC }
To validate the proposed methods, we compare the efficiency and accuracy of the FDD and FTC approaches with two state-of-the-art (SOTA) algorithms: the uniform passive FTC method \cite{Ke2023Uniform}, which operates without FDD, and the fixed-time FDD-based active FTC method \cite{yu2023integrated}. Since \cite{yu2023integrated} is not open-source, we combine the fixed-time FDD with NMPC to form one comparison. The benchmark methods are renamed as UnifP \cite{Ke2023Uniform} and FixedA \cite{yu2023integrated}. Four groups of tests are arranged to verify the robustness of the proposed method under motor or propeller failures, as sketched in Fig. \ref{sim_fdd_symbol}. In Test \#1 and Test \#2, the propeller failure and motor failure are injected during flying, which are involved during the takeoff in Test \#3 and Test \#4, respectively.

\subsubsection{FTC during Flight}

In Tests \#1 and \#2, we evaluate the FTC performance with the lemniscate trajectory at a speed of 2.0 $m/s$ with FixedA \cite{yu2023integrated} and our method. \textcolor{black}{The complete failure  of Propeller \#0 takes place during hovering or trajectory tracking, where each group contains 20 flight tests.} The specific results are shown in Table \ref{table_tracking}. Fig. \ref{sim_perpeller_ejection}  shows visualization with the proposed method in Test \#1, where the propeller-ejection failure occurs during quadrotor hovering. Fig. \ref{sim_fdd_propeller}(A) shows the flight trajectory visualizations with the proposed method in Test \#1 during quadrotor trajectory tracking. In Fig. \ref{sim_fdd_propeller}(B), the thrust-loss-based estimation method identifies an abnormal thrust command from rotor 0, which exceeds the critical threshold of $80\%$, enabling fault diagnosis to be completed within 0.184 $s$.  In contrast, the FDD time using FixedA \cite{yu2023integrated} is 0.253 $s$.  Fig. \ref{sim_fdd_propeller}(C) illustrates the state response during trajectory tracking. After a failure, the quadrotor requires nearly 3 $s$ to restore stability in roll and pitch attitudes, while entering a high-speed yaw spin mode with an angular velocity of 11.0 $rad/s$. In Test \#2, the motor-speed-based FDD method detects an abnormal motor speed, completing fault diagnosis within 0.028 $s$, while the FDD time using \cite{yu2023integrated} is 0.225 $s$. The success rates (SucR) for our method in flying phase of Tests \#1 and  \#2 are 90\% and 100\%, higher than those of 60\% and 70\% achieved by FixedA \cite{yu2023integrated}. \textcolor{black}{Additionally, the  false alarm rate (FAR) and missed detection rate (MDR) of both methods are 0\%.}  These findings highlight that designing specific FDD methods tailored to distinct types of rotor failures significantly enhances FDD efficiency, thereby improving SucR and flight safety.

\subsubsection{FTC during Takeoff}

We conduct Tests \#3 and \#4 to evaluate rotor failures  of Motor \#0 during takeoff, and the results are presented in Table \ref{table_hovering}. The quadrotor's acceleration and angular acceleration measurements reach the predefined thresholds, enabling fault detection to be completed within 0.020 $s$. The FDD times for FixedA \cite{yu2023integrated} under propeller and motor failures are 0.195 $s$ and 0.190 $s$, respectively. The minimum altitudes (MiniA) for \cite{Ke2023Uniform}, \cite{yu2023integrated}, and the proposed method are -0.262 $m$, -9.278 $m$, and 0 $m$, respectively. The results demonstrate that the takeoff attempts by \cite{yu2023integrated, Ke2023Uniform} result in crashes, whereas the proposed method ensures a successful and stable takeoff despite rotor failures.  Similar results are repeated in Test \#4. Compared to trajectory tracking flights in Tests \#1 and \#2, the allowable time for FDD during takeoff is more stringent, necessitating high-efficiency FDD and FTC methods specifically designed for this flight stage.  

\begin{table}[ht]
	\centering
	\caption{Performance Comparison of the FTC System during Trajectory Tracking. }
	\label{table_tracking}
	\begin{tabular}{@{}llcccc@{}}
		\toprule
		\text{} & \text{Methods}  & \text{FixedA \cite{yu2023integrated}} & \text{Our method}  \\

				\multirow{4}{*}{ \makecell{ Test \#1\\ Hovering \\ Propeller failure }} 
				 & \text{FDD time ($s$)} & 0.216 & $\bm{0.105}$ \\
				 & \text{SucR (\%)} & 100.0 & 100.0 \\
				& \text{MDR $ (\%)$}  & 0.0 & 0.0 \\
				& \text{FAR $ (\%)$}  & 0.0 & 0.0 \\
		
			\midrule
		\multirow{4}{*}{ \makecell{Test \#2\\ Hovering\\  Motor failure }}  
		& \text{FDD time ($s$)} & 0.207 & $\bm{0.032}$ \\
		& \text{SucR (\%)} & 100.0 & 100.0 \\
		& \text{MDR $ (\%)$}  & 0.0 & 0.0 \\
		& \text{FAR $ (\%)$}  & 0.0 & 0.0 \\
		
		\midrule
		\multirow{4}{*}{ \makecell{Test \#1\\ Trajectory tracking \\Propeller failure }}  
		& \text{FDD time ($s$)} &  0.253& $\bm{0.184}$ \\
		& \text{SucR (\%)} & 60.0 & $\bm{90.0}$ \\
		& \text{MDR $ (\%)$}  & 0.0 & 0.0 \\
		& \text{FAR $ (\%)$}  & 0.0 & 0.0 \\

		\midrule
		\multirow{4}{*}{ \makecell{Test \#2 \\ Trajectory tracking \\ Motor failure }}  
		& \text{FDD time ($s$)} & 0.225 & $\bm{ 0.028}$ \\
		& \text{SucR (\%)} & 70.0 & $\bm{100.0}$ \\
		& \text{MDR $ (\%)$}  & 0.0 & 0.0 \\
		& \text{FAR $ (\%)$}  & 0.0 & 0.0 \\

		\bottomrule
	\end{tabular}
\end{table}

\begin{table}[ht]
	\centering
	\caption{Performance Comparison of the FTC System during Hovering.  (Success: \checkmark ; Fail:$\times$ )}
	\label{table_hovering}
	\begin{tabular}{@{}llcccc@{}}
		\toprule
		\text{} & \text{Methods} & \text{UnifP \cite{Ke2023Uniform}} & \text{FixedA \cite{yu2023integrated}} & \text{Our method}  \\

		\midrule
		\multirow{4}{*}{\makecell{Test \#3 \\ Takeoff }} & \text{FDD time ($s$)} & /      & 0.195 & $\bm{0.020}$  \\
	    & \text{MiniA ($m$)}                              & -0.262 & -9.287 & $\bm{0}$  \\
    	& \text{Result}                                   & $\times$ & $\times$ & $\bm{\checkmark}$  \\	
    	
    	\midrule
    	\multirow{4}{*}{\makecell{Test \#4 \\ Takeoff  }} & \text{FDD time ($s$)} & /        & 0.190 & $\bm{0.020}$ \\
    	& \text{MiniA ($m$)}                              & -0.228   & -8.624 & $\bm{0}$  \\
    	& \text{Result}                                   & $\times$ & $\times$ & $\bm{\checkmark}$ \\	
		
		\bottomrule
	\end{tabular}
\end{table}

\textcolor{black}{
	We evaluate the computational performance of the NMPC on NVIDIA Jetson Orin NX  at a frequency of  200 Hz. The mean computational time is 1.88 $ms$, while the maximum time is 3.17 $ms$.  The result confirms that this platform provides sufficient computational capability for the real-time operation of NMPC in quadrotor control. However, achieving such high rates on lower-power platforms such as Raspberry Pi is challenging due to their limited computation capability. In such cases, it is necessary to either lower the NMPC frequency or to develop a more efficient solver for the NMPC optimization problem in order to ensure feasible real-time operation.
}

\subsection{Experiments on FTC System }
\label{FDD and FTC under Different Failure}

We conduct trajectory tracking tests in closed-loop flight experiments to validate the proposed algorithm, encompassing both fault detection and control reconfiguration. Rotor failures are induced in the quadrotor by code-triggered motor-stopping failure, hitting and unloading the propeller.

\begin{figure}
	\centering
	\includegraphics[width=0.950\linewidth]{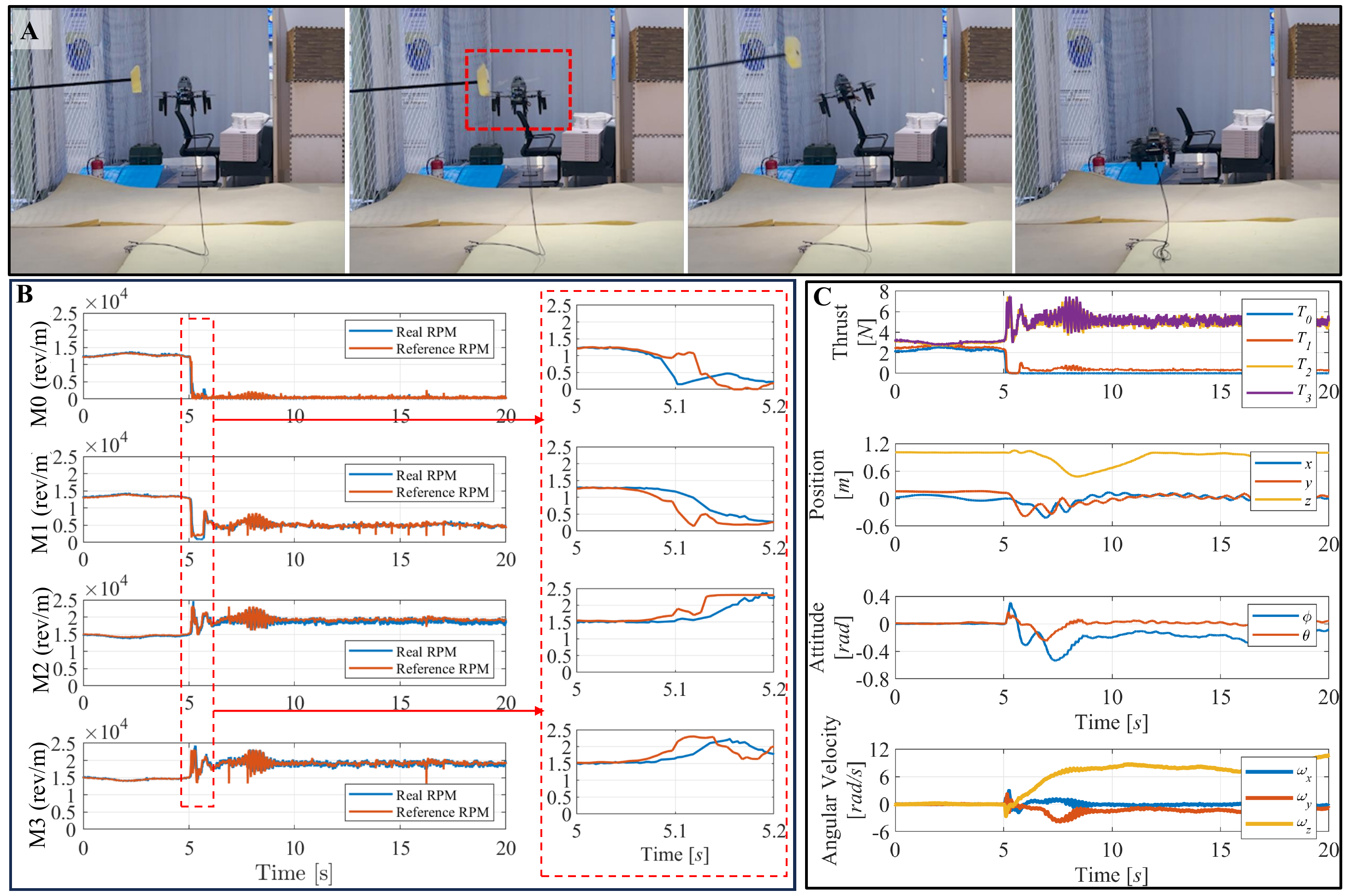}
	\caption{\textcolor{black}{Experimental results of the proposed method under motor failure. (A) Snapshots of inducing motor-stopping failure by striking the propeller with a wooden stick. (B) Comparison of the reference RPM and real-time measured RPM for all four motors. (C) Time history of the thrust and state responses controlled by the proposed FTC system.}}
	\label{hit_rpm_state}
\end{figure}

\begin{figure}
	\centering
	\includegraphics[width=0.950\linewidth]{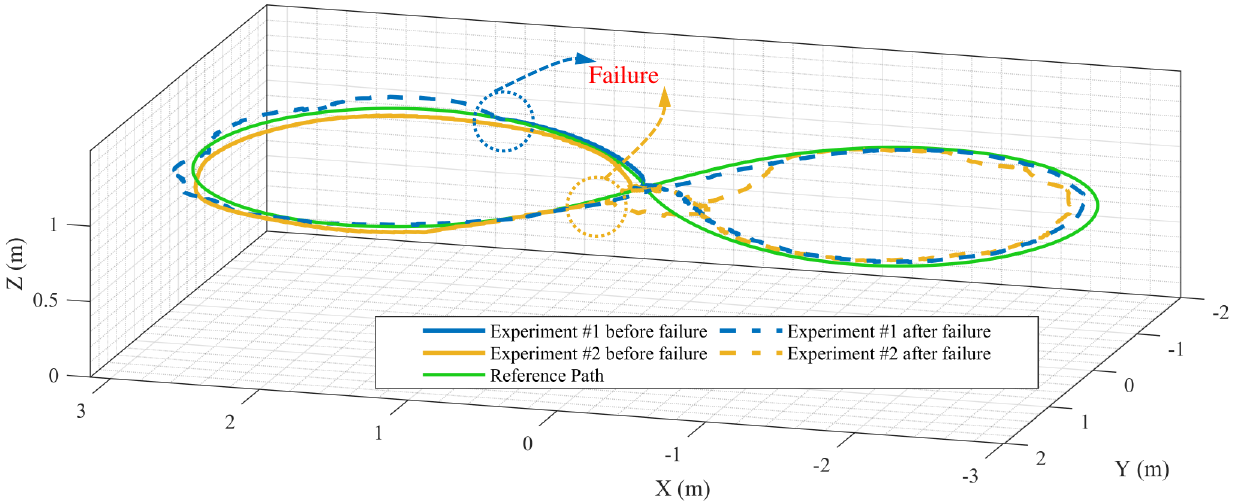}
	\caption{\textcolor{black}{Experimental results of trajectory tracking against motor-stopping failure. }}
	\label{exp_stop_motor}
\end{figure}

As illustrated in Fig. \ref{hit_rpm_state} (A), an experiment operator strikes the propellers of the hovering quadrotor directly with a wooden stick, simulating a collision that physically jams the motor temporarily. The quadrotor’s propeller is slicing through a sponge, and the quadrotor momentarily tilts due to the external collision force and motor failure. This motor stops during the collision, triggering the FDD mechanism for motor failure. Fig. \ref{hit_rpm_state} (B) shows the expected and actual RPMs of the four motors. \textcolor{black}{At $t = 5.102$ $s$, the actual and expected RPMs are 10185 $rev/m$ and 1528 $rev/m$, respectively.}  Additionally, since normal motor speed tracking is observed at $t = 5.075$ $s$, it can be inferred that the FDD completes within 27 $ms$. As depicted in Fig. \ref{hit_rpm_state} (C), once FDD is completed and the control system transitions to FTC mode, the speeds of the two motors adjacent to the faulty motor rapidly increase to counteract gravity, while the opposite motor to the faulty motor generates minimal thrust to maintain stability. The quadrotor’s altitude initially drops from hovering height, 1.0 $m$ to 0.45 $m$ but subsequently recovers to 1.0 $m$ while undergoing 11.0 $rad/s$ rotation. Furthermore, two experiments of motor-stopping failure are conducted by code-trigger way when the quadrotor is tracking a lemniscate trajectory at a speed of 1.0 $m/s$ as shown in Fig. \ref{exp_stop_motor}. The maximum tracking errors of the two experiments are 0.32 $m$ and 0.34 $m$, while the RMSEs are  0.15 $m$ and 0.15 $m$, respectively.

\begin{figure}
	\centering
	\includegraphics[width=0.950\linewidth]{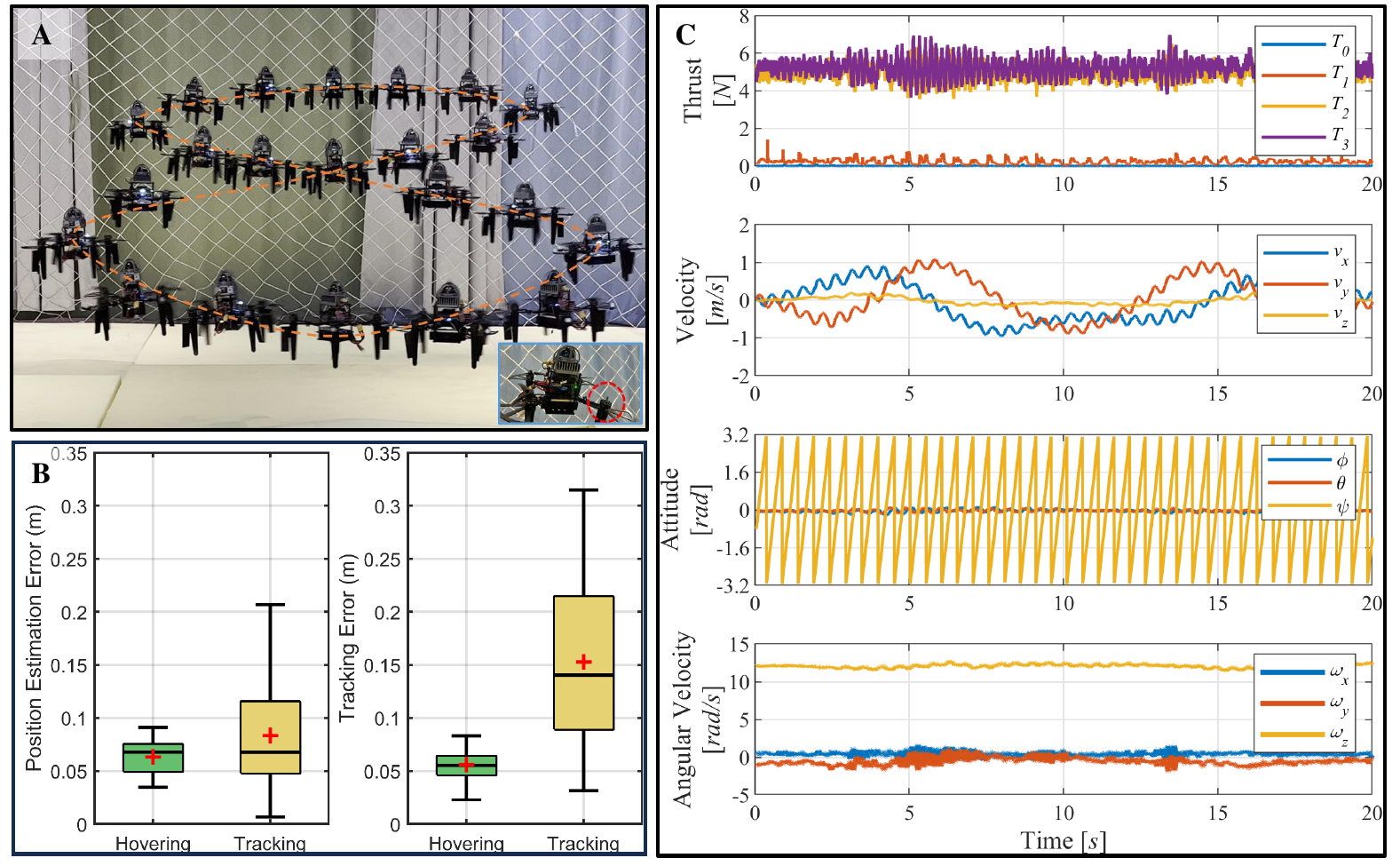}
	\caption{\textcolor{black}{FDD and FTC results of the proposed method against propeller-unloading failure. (A) Autonomous flight following a lemniscate trajectory despite the removal of a propeller. (B) Position estimation and trajectory tracking errors. (C) Time history of thrust and state responses controlled by the proposed FTC system.}}
	\label{remove_rotor_state}
\end{figure}

We also evaluate trajectory tracking performance by actively unloading a propeller, as illustrated in Fig. \ref{remove_rotor_state}(A). The lemniscate trajectory spans dimensions of 6.0 $m$ in the \(x\)-direction, 3.0 $m$ in the \(y\)-direction, and 1.0 $m$ in the \(z\)-direction  with maximum velocity 1.0 $m/s$. Fig. \ref{remove_rotor_state}(B) presents the RMSE for position estimation and tracking, where the MCS provides ground truth positions \cite{nokov2022}. The position estimation RMSEs during hovering and trajectory tracking are both 0.07 $m$. The tracking RMSE during hovering is  0.06 $m$, which increases to 0.14 $m$ during trajectory tracking. Despite high-speed rotation, the quadrotor demonstrates reliable tracking performance and stability in the roll and pitch channels, as shown in Fig. \ref{remove_rotor_state}(C). \textcolor{black}{In addition, although the proposed system shows strong performance in both simulations and real-world experiments, challenges from unmodeled dynamics, including aerodynamic uncertainties, environmental disturbances, and sensor degradation, can further impact post-failure quadrotor flight. Incorporating robust or adaptive components into the control and planning framework will be a valuable future research direction.}

\subsection{Autonomous Indoor Navigation}
\label{Autonomous navigation in dynamic and low-light environments}
\begin{figure}
	\centering
	\includegraphics[width=0.950\linewidth]{"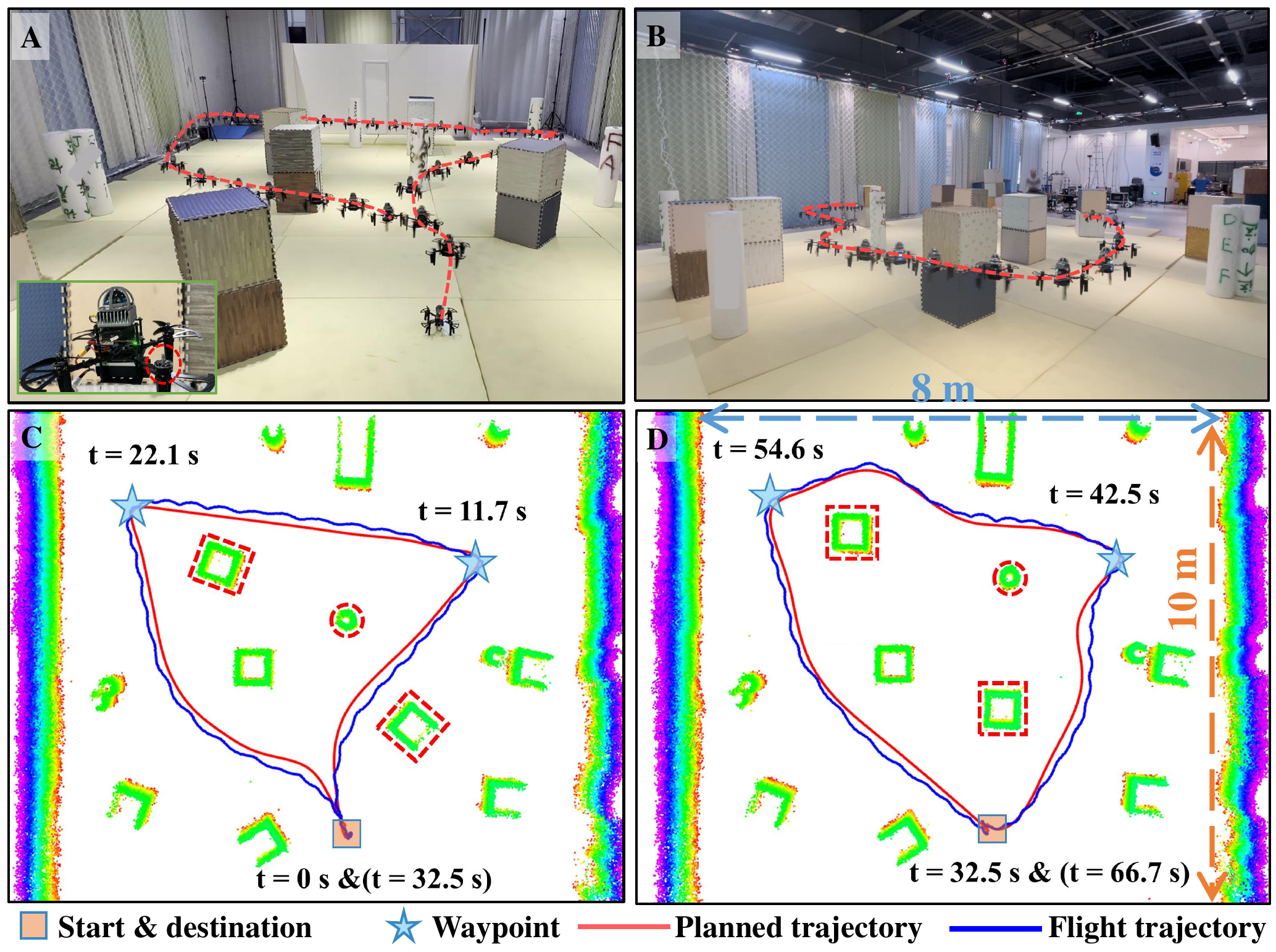"}
	\caption{\textcolor{black}{Autonomous waypoint navigation in an indoor environment with surrounding obstacles against propeller-unloading failure. (A) and (B) display front-view and side-view snapshots from the real-world experiment. (C) and (D) depict the first and second rounds of flight trajectory visualizations within the 3D point cloud map of the waypoint navigation experiments.  }}
	\label{indoornav}
\end{figure}

To fully evaluate the autonomous navigation capability of the proposed planning method, we conduct waypoint navigation experiments in indoor environments. The quadrotor, with one propeller unloaded, undergoes two types of tests: 1) The quadrotor navigates through 6 predefined waypoints while encountering manual-moving obstacles to assess its obstacle avoidance performance in complex environments (see Fig. \ref{indoornav}). 2) The quadrotor's autonomous navigation capability is further tested under low-light conditions and cluttered obstacles to evaluate its performance in a dark environment.\footnote{\url{https://sojustfish.github.io/Rotor-Failure-Aware/}}
 The total flight time is 66.7 $s$ in Fig. \ref{indoornav}, during which the quadrotor operates autonomously with maximum velocity 1.0 $m/s$. With no prior knowledge of the environment apart from the 6 predefined waypoints, the planner efficiently calculates smooth trajectories, enabling the quadrotor autonomous navigation despite an operator moving the obstacles (see the marked obstacles in Fig. \ref{indoornav}). The RMSE in flight is  0.08 $m$ and the maximum rotation speed is 10.7 $rad/s$. The computation times of the navigation modules running on the  NVIDIA Jetson Orin NX are also analyzed. The average processing time for the LiDAR inertial odometry and the trajectory planner is less than 20 $ms$.

\subsection{Autonomous Flight in the Wild}
\label{Autonomous navigation in Dense Garden}

We conduct outdoor flight experiments in an unknown forest,  covering an area of around 30 by 20 $m^{2}$ as illustrated in Fig. \ref{outdoor_nav}. Compared to the indoor experiments, the outdoor tests present additional challenges. The forest features trees of varying thicknesses and diverse natural vegetation. The passage, with a minimum width of less than 1.0 $m$, requires the quadrotor to adjust its altitude and position to navigate the cluttered environment. Wind disturbances with varying speeds and directions cause branches and leaves to sway, adversely affecting map, plan, and control stability.  
As shown in Fig. \ref{outdoor_nav}, the quadrotor successfully navigates through a dense area of trees with velocity 0.8 $m/s$, where the RMSE in flight is  0.09 $m$ and the maximum rotation speed is 12.6 $rad/s$. The results highlight the robustness of the proposed navigation system in handling outdoor environments.

\section{Conclusion }
\label{Conclusion}

In this study, we underscore the importance of ensuring autonomous navigation and flight stability for a quadrotor experiencing rotor failures in unknown environments. To meet these critical requirements, we design a composite FDD-based FTC system that actively and promptly acquires information on both motor and propeller failures, while mitigating vibrations caused by rotor imbalance from takeoff to trajectory tracking. Furthermore, we propose a rotor-failure-aware planning algorithm that formulates an optimization problem incorporating quadrotor dynamic feasibility and obstacle avoidance. Our approach systematically addresses the challenges of achieving collision-free flight in challenging environments for quadrotors with rotor failures. Comprehensive simulations and experimental validations demonstrate the practicality and effectiveness of the proposed method. \textcolor{black}{Nevertheless, the current system has limitations. In particular, scenarios involving dual rotor loss present severe challenges to controllability that are not yet addressed in this work. Furthermore, strong wind conditions can introduce unpredictable disturbances that degrade control performance. Addressing these scenarios will be an important direction for future research.}

\ifCLASSOPTIONcaptionsoff
\newpage
\fi

\bibliographystyle{IEEEtran}
\bibliography{IEEEabrv,mylib}

\begin{thebibliography}{10}
\providecommand{\url}[1]{#1}
\csname url@samestyle\endcsname
\providecommand{\newblock}{\relax}
\providecommand{\bibinfo}[2]{#2}
\providecommand{\BIBentrySTDinterwordspacing}{\spaceskip=0pt\relax}
\providecommand{\BIBentryALTinterwordstretchfactor}{4}
\providecommand{\BIBentryALTinterwordspacing}{\spaceskip=\fontdimen2\font plus
\BIBentryALTinterwordstretchfactor\fontdimen3\font minus
  \fontdimen4\font\relax}
\providecommand{\BIBforeignlanguage}[2]{{%
\expandafter\ifx\csname l@#1\endcsname\relax
\typeout{** WARNING: IEEEtran.bst: No hyphenation pattern has been}%
\typeout{** loaded for the language `#1'. Using the pattern for}%
\typeout{** the default language instead.}%
\else
\language=\csname l@#1\endcsname
\fi
#2}}
\providecommand{\BIBdecl}{\relax}
\BIBdecl

\bibitem{zhou2022swarm}
X.~Zhou, X.~Y. Wen, Z.~P. Wang, Y.~M. Gao, H.~J. Li, Q.~Wang, T.~K. Yang, H.~J.
  Lu, Y.~J. Cao, C.~Xu \emph{et~al.}, ``Swarm of micro flying robots in the
  wild,'' \emph{Science Robotics}, vol.~7, no.~66, p. eabm5954, 2022.

\bibitem{Cui2024FastSim}
C.~Cui, X.~B. Zhou, M.~Wang, F.~Gao, and C.~Xu, ``Fastsim: A modular and
  plug-and-play simulator for aerial robots,'' \emph{IEEE Robotics and
  Automation Letters}, vol.~9, no.~6, pp. 5823--5830, 2024.

\bibitem{tabib2021autonomous}
W.~Tabib, K.~Goel, J.~Yao, C.~Boirum, and N.~Michael, ``Autonomous cave
  surveying with an aerial robot,'' \emph{IEEE Transactions on Robotics},
  vol.~38, no.~2, pp. 1016--1032, 2021.

\bibitem{witze2024first}
A.~Witze, ``First aircraft to fly on mars dies—but leaves a legacy of
  science,'' \emph{Nature}, vol. 626, no. 7998, pp. 244--244, 2024.

\bibitem{ackerman2019swiss}
E.~Ackerman, ``Swiss post suspends drone delivery service after second crash,''
  \emph{IEEE Spectrum}, July 2019.

\bibitem{Wu2021Nonlinear}
Y.~H. Wu, K.~J. Hu, X.~M. Sun, and Y.~H. Ma, ``Nonlinear control of quadrotor
  for fault tolerance: A total failure of one actuator,'' \emph{IEEE
  Transactions on Systems, Man, and Cybernetics: Systems}, vol.~51, no.~5, pp.
  2810--2820, 2021.

\bibitem{Stephan2018Linear}
J.~Stephan, L.~Schmitt, and W.~Fichter, ``Linear parameter-varying control for
  quadrotors in case of complete actuator loss,'' \emph{Journal of Guidance,
  Control, and Dynamics}, vol.~41, pp. 2232--2246, 2018.

\bibitem{Ke2023Uniform}
C.~X. Ke, K.~Y. Cai, and Q.~Quan, ``Uniform passive fault-tolerant control of a
  quadcopter with one, two, or three rotor failure,'' \emph{IEEE Transactions
  on Robotics}, vol.~39, no.~6, pp. 4297--4311, 2023.

\bibitem{zhang2008bibliographical}
Y.~M. Zhang and J.~Jiang, ``Bibliographical review on reconfigurable
  fault-tolerant control systems,'' \emph{Annual Reviews in Control}, vol.~32,
  no.~2, pp. 229--252, 2008.

\bibitem{lippiello2014emergency}
V.~Lippiello, F.~Ruggiero, and D.~Serra, ``Emergency landing for a quadrotor in
  case of a propeller failure: A {PID} based approach,'' in \emph{Proceedings
  of IEEE International Symposium on Safety, Security, and Rescue Robotics},
  2014, pp. 4782--4788.

\bibitem{Merheb2017Emergency}
A.~Merheb, H.~Noura, and F.~Bateman, ``Emergency control of {AR} drone
  quadrotor {UAV} suffering a total loss of one rotor,'' \emph{IEEE/ASME
  Transactions on Mechatronics}, vol.~22, no.~2, pp. 961--971, 2017.

\bibitem{yu2023integrated}
X.~Yu, S.~C. Zhou, K.~X. Guo, Y.~M. Zhang, and L.~Guo, ``Integrated
  reconfiguration mechanism for quadrotors with capability analysis against
  rotor failure,'' \emph{Journal of Guidance, Control, and Dynamics}, vol.~46,
  no.~2, pp. 401--409, 2023.

\bibitem{sun2020incremental}
S.~H. Sun, X.~R. Wang, Q.~P. Chu, and C.~de~Visser, ``Incremental nonlinear
  fault-tolerant control of a quadrotor with complete loss of two opposing
  rotors,'' \emph{IEEE Transactions on Robotics}, vol.~37, no.~1, pp. 116--130,
  2020.

\bibitem{Sun2021Autonomous}
S.~H. Sun, G.~Cioffi, C.~de~Visser, and D.~Scaramuzza, ``Autonomous quadrotor
  flight despite rotor failure with onboard vision sensors: Frames vs.
  events,'' \emph{IEEE Robotics and Automation Letters}, vol.~6, no.~2, pp.
  580--587, 2021.

\bibitem{Chenxu}
C.~X. Ke, K.~Y. Cai, and Q.~Quan, ``Uniform fault-tolerant control of a
  quadcopter with rotor failure,'' \emph{IEEE/ASME Transactions on
  Mechatronics}, vol.~28, no.~1, pp. 507--517, 2023.

\bibitem{Mueller2014Stability}
M.~W. {Mueller} and R.~{D'Andrea}, ``Stability and control of a quadrocopter
  despite the complete loss of one, two, or three propellers,'' in
  \emph{Proceedings of the IEEE International Conference on Robotics and
  Automation}, 2014, pp. 45--52.

\bibitem{liu2024audio}
W.~S. Liu, C.~Liu, S.~Sajedi, H.~Su, X.~Liang, and M.~H. Zheng, ``An
  audio-based risky flight detection framework for quadrotors,'' \emph{IET
  Cyber-Systems and Robotics}, vol.~6, no.~1, p. e12105, 2024.

\bibitem{o2024learning}
M.~O'Connell, J.~Cho, M.~Anderson, and S.~J. Chung, ``Learning-based
  minimally-sensed fault-tolerant adaptive flight control,'' \emph{IEEE
  Robotics and Automation Letters}, vol.~9, no.~6, pp. 5198--5205, 2024.

\bibitem{mao2024propeller}
J.~Mao, J.~Yeom, S.~Nair, and G.~Loianno, ``From propeller damage estimation
  and adaptation to fault tolerant control: Enhancing quadrotor resilience,''
  \emph{IEEE Robotics and Automation Letters}, vol.~9, no.~5, pp. 4297--4304,
  2024.

\bibitem{nam2024fault}
Y.~Nam, K.~Lee, H.~S. Shin, and C.~Kwon, ``Fault tolerant motion planning for a
  quadrotor subject to complete rotor failure,'' \emph{Aerospace Science and
  Technology}, vol. 154, p. 109529, 2024.

\bibitem{zhou2024internal}
X.~B. Zhou, M.~Wang, C.~Cui, Y.~C. Wang, C.~Xu, and F.~Gao, ``Internal and
  external disturbances aware motion planning and control for quadrotors,''
  \emph{IET Cyber-Systems and Robotics}, vol.~6, no.~3, p. e12122, 2024.

\bibitem{Liu2023Simultaneous}
Z.~Liu and L.~L. Cai, ``Simultaneous planning and execution for safe flight of
  quadrotors suffering one rotor loss and disturbance,'' \emph{IEEE
  Transactions on Aerospace and Electronic Systems}, vol.~59, no.~5, pp.
  5731--5747, 2023.

\bibitem{Flatness2012Chamseddine}
A.~Chamseddine, Y.~M. Zhang, C.~A. Rabbath, C.~Join, and D.~Theilliol,
  ``Flatness-based trajectory planning/replanning for a quadrotor unmanned
  aerial vehicle,'' \emph{IEEE Transactions on Aerospace and Electronic
  Systems}, vol.~48, no.~4, pp. 2832--2848, 2012.

\bibitem{Li2022AutoTrans}
H.~J. Li, H.~K. Wang, C.~Feng, F.~Gao, B.~Y. Zhou, and S.~J. Shen, ``Autotrans:
  A complete planning and control framework for autonomous {UAV} payload
  transportation,'' \emph{IEEE Robotics and Automation Letters}, vol.~8,
  no.~10, pp. 6859--6866, 2023.

\bibitem{wang2022neither}
L.~Q. Wang, H.~Xu, Y.~C. Zhang, and S.~J. Shen, ``Neither fast nor slow: How to
  fly through narrow tunnels,'' \emph{IEEE Robotics and Automation Letters},
  vol.~7, no.~2, pp. 5489--5496, 2022.

\bibitem{Wu2021External}
Y.~W. Wu, Z.~M. Ding, C.~Xu, and F.~Gao, ``External forces resilient safe
  motion planning for quadrotor,'' \emph{IEEE Robotics and Automation Letters},
  vol.~6, no.~4, pp. 8506--8513, 2021.

\bibitem{Shi2022Neural}
G.~Y. Shi, W.~Hönig, X.~C. Shi, Y.~S. Yue, and S.~J. Chung, ``Neural-swarm2:
  Planning and control of heterogeneous multirotor swarms using learned
  interactions,'' \emph{IEEE Transactions on Robotics}, vol.~38, no.~2, pp.
  1063--1079, 2022.

\bibitem{chen2023self}
N.~Chen, F.~Z. Kong, W.~Xu, Y.~X. Cai, H.~T. Li, D.~J. He, Y.~M. Qin, and
  F.~Zhang, ``A self-rotating, single-actuated {UAV} with extended sensor field
  of view for autonomous navigation,'' \emph{Science Robotics}, vol.~8, no.~76,
  p. eade4538, 2023.

\bibitem{Mavros2024git}
Mavros, https://github.com/mavlink/mavros.

\bibitem{xu2022fast}
W.~Xu, Y.~X. Cai, D.~J. He, J.~R. Lin, and F.~Zhang, ``Fast-lio2: Fast direct
  lidar-inertial odometry,'' \emph{IEEE Transactions on Robotics}, vol.~38,
  no.~4, pp. 2053--2073, 2022.

\bibitem{Brescianini2020Tilt}
D.~Brescianini and R.~D’Andrea, ``Tilt-prioritized quadrocopter attitude
  control,'' \emph{IEEE Transactions on Control Systems Technology}, vol.~28,
  no.~2, pp. 376--387, 2020.

\bibitem{zhou2019robust}
B.~Y. Zhou, F.~Gao, L.~Q. Wang, C.~H. Liu, and S.~J. Shen, ``Robust and
  efficient quadrotor trajectory generation for fast autonomous flight,''
  \emph{IEEE Robotics and Automation Letters}, vol.~4, no.~4, pp. 3529--3536,
  2019.

\bibitem{zhou2020ego}
X.~Zhou, Z.~P. Wang, H.~K. Ye, C.~Xu, and F.~Gao, ``Ego-planner: An {ESDF}-free
  gradient-based local planner for quadrotors,'' \emph{IEEE Robotics and
  Automation Letters}, vol.~6, no.~2, pp. 478--485, 2020.

\bibitem{ren2025safety}
Y.~Ren, F.~Zhu, G.~Lu, Y.~Cai, L.~Yin, F.~Kong, J.~Lin, N.~Chen, and F.~Zhang,
  ``Safety-assured high-speed navigation for {MAVs},'' \emph{Science Robotics},
  vol.~10, no.~98, p. eado6187, 2025.

\bibitem{liu1989limited}
D.~C. Liu and J.~Nocedal, ``On the limited memory {BFGS } method for large
  scale optimization,'' \emph{Mathematical programming}, vol.~45, no.~1, pp.
  503--528, 1989.

\bibitem{houska2011acado}
B.~Houska, H.~J. Ferreau, and M.~Diehl, ``Acado toolkit—{An} open-source
  framework for automatic control and dynamic optimization,'' \emph{Optimal
  control applications and methods}, vol.~32, no.~3, pp. 298--312, 2011.

\bibitem{ferreau2014qpoases}
H.~J. Ferreau, C.~Kirches, A.~Potschka, H.~G. Bock, and M.~Diehl, ``{QPOASES}:
  A parametric active-set algorithm for quadratic programming,''
  \emph{Mathematical Programming Computation}, vol.~6, pp. 327--363, 2014.

\bibitem{nokov2022}
\BIBentryALTinterwordspacing
NOKOV, ``Nokov optical motion capture system,'' 2022. [Online]. Available:
  \url{https://en.nokov.com/}
\BIBentrySTDinterwordspacing

\end{thebibliography}

\vspace{-1.6cm}

\begin{IEEEbiography}[{\includegraphics[width=1in,height=1.25in,clip,keepaspectratio]{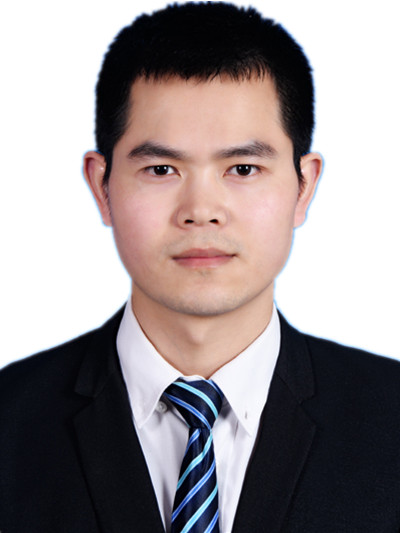}}]{Xiaobin Zhou}
 received the B.S. degree from Northeastern University, Shenyang, China, in 2015, and the Ph.D. degree from Hunan University, Changsha, China, in 2021, both in mechanical engineering. From 2018 to 2021, he was a Visiting Student with Concordia University, Montreal, QC, Canada, and Beihang University, Beijing, China. From 2021 to 2022, he was an Engineer with XPeng AeroHT. From 2022 to 2025, he was a Postdoctoral Fellow with Zhejiang University. He is currently an Assistant Professor with the School of Robotics and Automation, Nanjing University, China. His research interests include aerial robot design, autonomous navigation, and multi-robot systems.
\end{IEEEbiography}

\vspace{-1.2cm}

\begin{IEEEbiography}[{\includegraphics[width=1in,height=1.25in,clip,keepaspectratio]{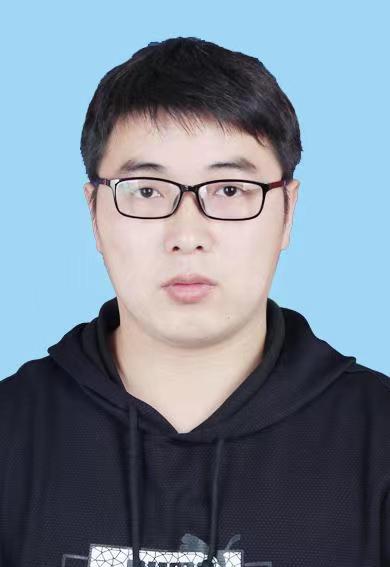}}]{Miao Wang}
 received his Ph.D. degree in Cyber Science and Engineering from Southeast University, China, in July 2023. He then conducted postdoctoral research at Zhejiang University, focusing on path planning and swarm control. He is currently an Associate Research Fellow at Nanjing University, China. His research interests include UAV structural design, trajectory/path planning, and swarm intelligence and control, aiming to improve the efficiency and reliability of unmanned aerial systems through innovative design and control strategies.
\end{IEEEbiography}

\vspace{-1.2cm}

\begin{IEEEbiography}[{\includegraphics[width=1in,height=1.25in,clip,keepaspectratio]{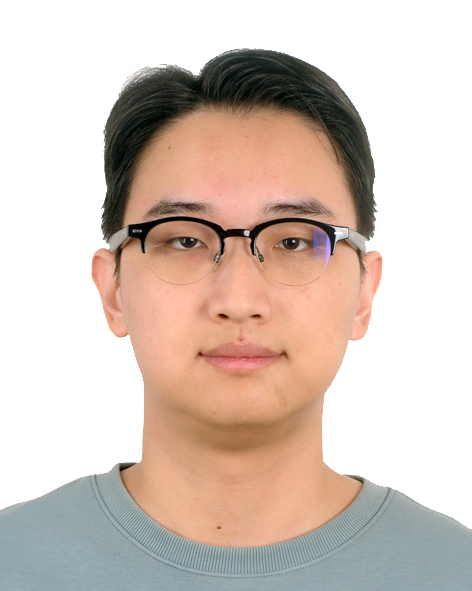}}]{Chengao Li}
  received the B.Eng. degree in Aerospace Engineering from the University of Nottingham Ningbo China in 2024. In the same year, he served as a Research Assistant at Zhejiang University. He is currently an MPhil student with the Department of Aeronautical and Aviation Engineering, The Hong Kong Polytechnic University, Hong Kong, China. His research interests include aerial robotics, autonomous systems, and aviation engineering.\textbf{}
\end{IEEEbiography}

\vspace{-1.2cm}

\begin{IEEEbiography}[{\includegraphics[width=1in,height=1.25in,clip,keepaspectratio]{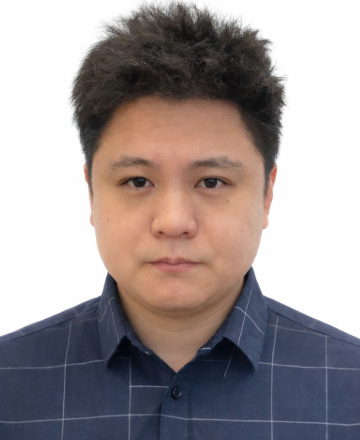}}]{Can Cui}
 received the B.S. degree in control science and engineering from Zhejiang University, Hangzhou, China, in 2012, and the MSc degree in computer science from The University of Hong Kong, Hong Kong, China, in 2014. He is a researcher at Huzhou Institute of Zhejiang University, Huzhou, a Ph.D student in the College of Control Science and Engineering, Zhejiang University, Hangzhou, and he is also the CEO of Ningbo Novation Technology Co. Ltd. His research interests include aerial robotics, autonomous navigation, and simulation systems.
\end{IEEEbiography}

\vspace{-0.2cm}

\begin{IEEEbiography}[{\includegraphics[width=1in,height=1.25in,clip,keepaspectratio]{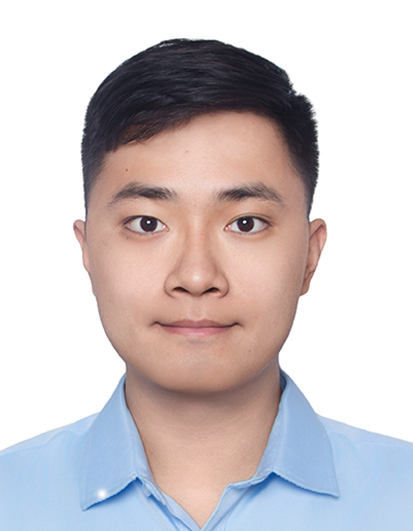}}]{Ruibin Zhang} 
 received the B.S. degree in mechatronics from Zhejiang University, Hangzhou, China, in 2021, where he is currently working toward the Ph.D. degree in automation under the supervision of Fei Gao, with the FAST Lab, Zhejiang University. His research interests include environmental perception, motion planning, and autonomous navigation for mobile robots.
\end{IEEEbiography}

\vspace{-1.2cm}

\begin{IEEEbiography}[{\includegraphics[width=1in,height=1.25in,clip,keepaspectratio]{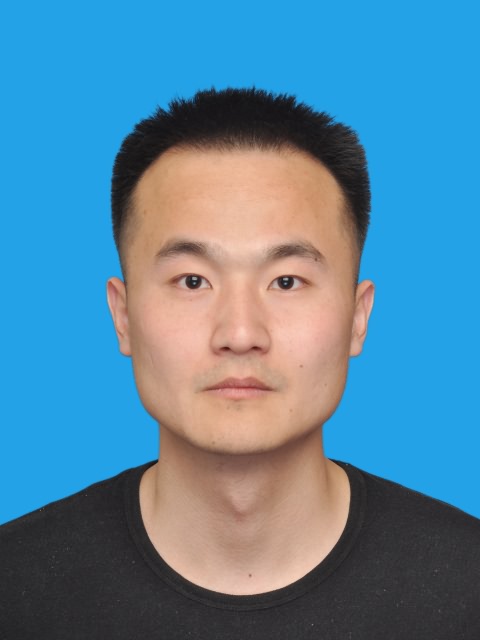}}]{Yongchao Wang} 
 received the Ph.D. degree from Beihang University in 2021. He is a post-doctoral fellow with the School of Aeronautic Science and Engineering, Beihang University. His research interests include the trajectory planning and control for the multiple quadcopter aerial transportation system, multi-UAV cooperative decision making, and reinforcement learning.
	
\end{IEEEbiography}

\vspace{-1.2cm}

\begin{IEEEbiography}[{\includegraphics[width=1in,height=1.25in,clip,keepaspectratio]{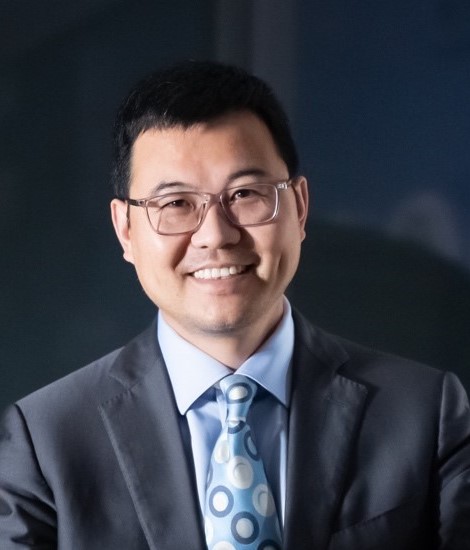}}]{Chao Xu}
	received the Ph.D. degree in Mechanical Engineering from Lehigh University in 2010. He is currently Associate Dean and Professor at the College of Control Science and Engineering, Zhejiang University. He serves as the inaugural Dean of ZJU Huzhou Institute, as well as plays the role of the Managing Editor for \textit{IET Cyber-Systems \& Robotics}. 
	His research expertise is Flying Robotics, Control-theoretic Learning. Prof. Xu has published over 100 papers in international journals, including \textit{Science Robotics}, \textit{Nature Machine Intelligence}, etc.
\end{IEEEbiography}

\vspace{-1.2cm}

\begin{IEEEbiography}[{\includegraphics[width=1in,height=1.25in,clip,keepaspectratio]{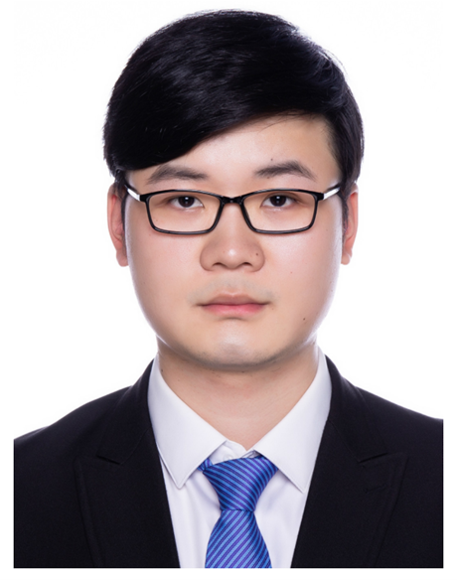}}]{Fei Gao}
	received the Ph.D. degree in electronic and computer engineering from the Hong Kong University of Science and Technology,Hong Kong, in 2019. 
	He is currently a tenured associate professor at the Department of Control Science and Engineering, Zhejiang University. 
	His research interests include aerial robots, autonomous navigation,motion planning, optimization, and localization and mapping. 
	He is also the founder of Differential Robotics Ltd. 
\end{IEEEbiography}

\end{document}